\documentclass[10pt,twocolumn,letterpaper]{article}

\usepackage{cvpr}              %

\usepackage{amsmath}
\usepackage{amssymb}

\usepackage{amsmath,amsfonts,bm}

\def\eqref#1{equation~\ref{#1}}

\def\1{\bm{1}}
\newcommand{\train}{\mathcal{D}}

\def\rvc{{\mathbf{c}}}

\def\rvh{{\mathbf{h}}}

\def\rvp{{\mathbf{p}}}

\def\rvs{{\mathbf{s}}}

\def\rvx{{\mathbf{x}}}
\def\rvy{{\mathbf{y}}}

\DeclareMathAlphabet{\mathsfit}{\encodingdefault}{\sfdefault}{m}{sl}
\SetMathAlphabet{\mathsfit}{bold}{\encodingdefault}{\sfdefault}{bx}{n}

\DeclareMathOperator{\Tr}{Tr}

\usepackage{url}
\usepackage{booktabs}
\usepackage{graphicx}
\usepackage{multirow}
\usepackage[table]{xcolor}
\usepackage{color, colortbl}
\usepackage{algorithm}
\usepackage{listings}
\usepackage{lstautogobble}
\usepackage[accsupp]{axessibility}

\definecolor{bluecolor}{HTML}{0071bc}
\usepackage[pagebackref,breaklinks,colorlinks,allcolors=bluecolor]{hyperref}

\usepackage[capitalize]{cleveref}
\crefname{section}{Sec.}{Secs.}
\Crefname{section}{Section}{Sections}
\Crefname{table}{Table}{Tables}
\crefname{table}{Tab.}{Tabs.}

\def \eg {{e.g.,}\xspace}
\def \ie {{i.e.,}\xspace}

\newcommand*\samethanks[1][\value{footnote}]{\footnotemark[#1]}

\definecolor{azure(colorwheel)}{rgb}{0.0, 0.5, 1.0}

\def\cvprPaperID{9110} %
\def\confName{CVPR}
\def\confYear{2023}

\begin{document}

\title{Semi-supervised learning made simple with self-supervised clustering}

\author{Enrico Fini\thanks{\scriptsize{Enrico Fini and Pietro Astolfi contributed equally.}} \,$^{\textcolor{azure(colorwheel)}{1}}$
\quad Pietro Astolfi\samethanks \, $^{\textcolor{azure(colorwheel)}{1,2}}$
\quad Karteek Alahari$^{\textcolor{azure(colorwheel)}{2}}$
\quad Xavier Alameda-Pineda$^{\textcolor{azure(colorwheel)}{2}}$\\
\quad Julien Mairal$^{\textcolor{azure(colorwheel)}{2}}$\vspace{5px}
\quad Moin Nabi$^{\textcolor{azure(colorwheel)}{3}}$
\quad Elisa Ricci$^{\textcolor{azure(colorwheel)}{1,4}}$\\
\normalsize{$^{\textcolor{azure(colorwheel)}{1}}$ University of Trento
\quad $^{\textcolor{azure(colorwheel)}{2}}$ Inria\thanks{\scriptsize{Univ. Grenoble Alpes, CNRS, Grenoble INP, LJK, 38000 Grenoble, France.}}
\quad $^{\textcolor{azure(colorwheel)}{3}}$ SAP AI Research
\quad $^{\textcolor{azure(colorwheel)}{4}}$ Fondazione Bruno Kessler}
}
\maketitle

\begin{abstract}
  Self-supervised learning models have been shown to learn rich visual representations without requiring human annotations. However, in many real-world scenarios, labels are partially available, motivating a recent line of work on semi-supervised methods inspired by self-supervised principles. In this paper, we propose a conceptually simple yet empirically powerful approach to turn clustering-based self-supervised methods such as SwAV or DINO into semi-supervised learners. More precisely, we introduce a multi-task framework merging a supervised objective using ground-truth labels and a self-supervised objective relying on clustering assignments with a single cross-entropy loss. This approach may be interpreted as imposing the cluster centroids to be class prototypes.  Despite its simplicity, we provide empirical evidence that our approach is highly effective and achieves state-of-the-art performance on CIFAR100 and ImageNet. %
\end{abstract}

\section{Introduction}
\label{sec:intro}
In recent years, self-supervised learning became the dominant paradigm for unsupervised visual representation learning. In particular, much experimental evidence shows that augmentation-based self-supervision~\cite{bachman2019learning, chen2020simple, he2020momentum, chen2021exploring, grill2020bootstrap, ermolov2021whitening, zbontar2021barlow, bardes2022vicreg, bardes2022vicregl, asano2020self, caron2018deep, caron2020unsupervised, caron2021emerging} can produce powerful representations of unlabeled data. Such models, although trained without supervision, can be naturally used for supervised downstream tasks via simple fine-tuning. However, the most suitable way to leverage self-supervision is perhaps by multi-tasking the self-supervised objective with a custom (possibly supervised) objective. Based on this idea, the community has worked on re-purposing self-supervised methods in other sub-fields of computer vision, as for instance in domain adaptation \cite{da2022dual}, novel class discovery~\cite{fini2021unified, zhong2021neighborhood}, continual learning~\cite{fini2022self} and semi-supervised learning~\cite{zhai2019s4l, assran2020supervision, chen2020big, cai2021exponential}.

\begin{figure}
    \centering
    \includegraphics[width=0.95\columnwidth]{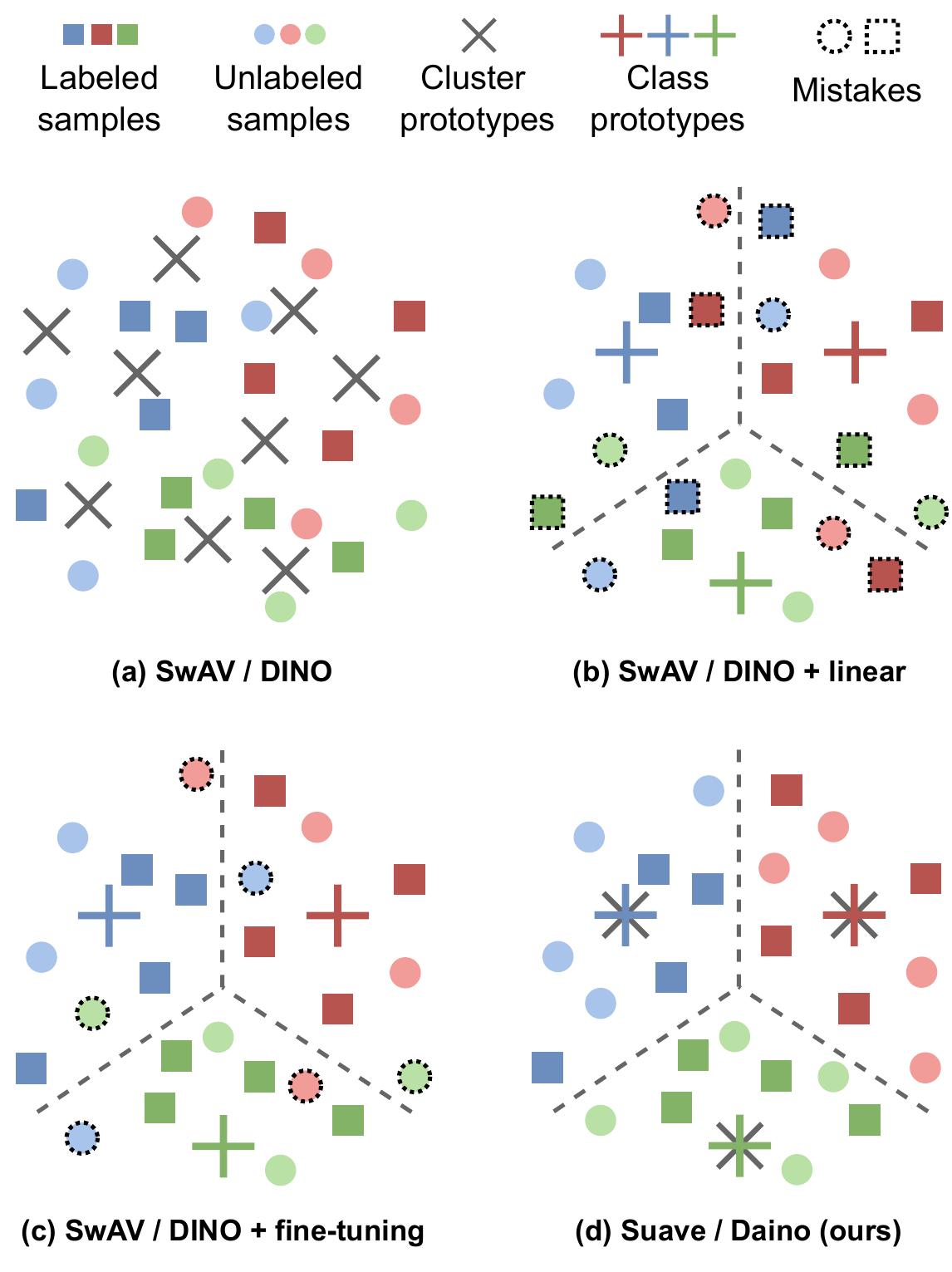}
    \vspace{-1mm}
    \caption{Schematic illustration of the motivation behind the proposed semi-supervised framework. (a) Self-supervised clustering methods like SwAV \cite{caron2020unsupervised} and DINO \cite{caron2021emerging} compute cluster prototypes that are not necessarily well aligned with semantic categories, but they do not require labeled data. (b) Adding a linear classifier provides class prototypes, but the labeled (and unlabeled) samples are not always correctly separated. (c) Fine-tuning can help separating labeled data. (d) Our framework learns cluster prototypes that are aligned with class prototypes thus correctly separating both labeled and unlabeled data.
    }
    \label{fig:teaser}
    \vspace{-4mm}
\end{figure}

One of the areas that potentially benefits from the advancements in unsupervised representation learning is semi-supervised learning. This is mainly due to the fact that in semi-supervised learning it is crucial to efficiently extract the information in the unlabeled set to improve the classification accuracy on the labeled classes. Indeed, several powerful semi-supervised methods \cite{zhai2019s4l, wang2020enaet,cai2021exponential,assran2021semi} were built upon this idea. 

In the self-supervised learning landscape, arguably the most successful methods belong to the clustering-based family, such as DeepCluster v2 \cite{caron2018deep}, SwAV \cite{caron2020unsupervised} and DINO \cite{caron2021emerging}. These methods learn representations by contrasting predicted cluster assignments of correlated views of the same image. To avoid collapsed solutions and group samples together, they use simple clustering-based pseudo-labeling algorithms such as {k-means} and Sinkhorn-Knopp to generate the assignments. A peculiar fact about this family of methods is the discretization of the feature space, that allows them to use techniques that were originally developed for supervised learning. Indeed, similar to supervised learning, the cross-entropy loss is adopted to compare the assignments, as they represent probability distributions over the set of clusters. 

In this paper, we propose a new approach for semi-supervised learning based on the simple observation that clustering-based methods are amenable to be adapted to a semi-supervised learning setting: the cluster prototypes can be replaced with class prototypes learned with supervision and the same loss function can be used for both labeled and unlabeled data. In practice, semi-supervised learning can be achieved by multi-tasking the self-supervised and supervised objectives. This encourages the network to cluster unlabeled samples around the centroids of the classes in the feature space. By leveraging on these observations we propose a new framework for semi-supervised methods based on self-supervised clustering. We experiment with two instances of that framework: \textit{Suave} %
and \textit{Daino}, %
the semi-supervised counterparts of SwAV and DINO. These methods have several favorable properties: i) they are efficient at learning representations from unlabeled data since they are based on the top-performing self-supervised methods; ii) they extract relevant information for the semantic categories associated with the data thanks to the supervised conditioning; iii) they are easy to implement as they are based on the multi-tasking of two objectives.
The motivation behind our proposal is also illustrated in Fig.~\ref{fig:teaser}.
As shown in the figure, our multi-tasking approach enables to compute cluster centers that are aligned with class prototypes thus correctly separating both labeled and unlabeled data.

Our \textbf{contributions} can be summarized as follows:
\begin{itemize}
    \item We propose a new framework for semi-supervised learning based on the multi-tasking of a supervised objective on the labeled data and a clustering-based self-supervised objective on the unlabeled samples;
    \item We experiment with two representatives of such framework: Suave and Daino, semi-supervised extensions of SwAV \cite{caron2020unsupervised} and DINO \cite{caron2021emerging}. These methods, while simple to implement, are powerful and efficient semi-supervised learners;
    \item Our methods outperform state-of-the-art approaches, often relying on multiple \textit{ad hoc} components, both on common small scale (CIFAR100) and large scale (ImageNet) benchmarks, setting a new state-of-the-art on semi-supervised learning.
\end{itemize}\section{Related works}
\label{sec:rel_works}

\paragraph{Self-supervised learning.} The self-supervised literature has rapidly become extremely vast \cite{jaiswal2020survey}. Excluding a recent trend involving denoising autoencoders~\cite{vincent2010stacked} combined with vision transformers~\cite{dosovitskiy2021image} like in MAE~\cite{he2022masked}, the vast majority of existing methods are still relying on multiple views derived from each sample via data augmentation.
These augmentation-based methods ~\cite{bachman2019learning, chen2020simple, he2020momentum, chen2021exploring, grill2020bootstrap, ermolov2021whitening, zbontar2021barlow, bardes2022vicreg, bardes2022vicregl, asano2020self, caron2018deep, caron2020unsupervised, caron2021emerging} are capable to learn rich representations during unsupervised pre-training, achieving performance comparable to their supervised counterparts when fine-tuned with the labels. Basically, these methods enforce correlated views of the same input to have coherent representations in latent space such that the model becomes invariant to the augmentations applied. This corresponds to maximizing the mutual information between views' representations and, in the literature, it has been done using different loss functions.

Contrastive-based methods \cite{bachman2019learning, chen2020simple, he2020momentum, huynh2022boosting} define a loss based on noise-contrastive estimation~\cite{gutmann2010noise} as instance discrimination~\cite{wu2018unsupervised} between positive (correlated) and negative views (the remaining samples in the mini-batch). A major drawback of these methods is that they require large mini-batches to have representative negative samples. To overcome this difficulty, consistency-based methods like \cite{chen2021exploring, grill2020bootstrap} propose to maximize the cosine similarity of positive pairs without considering negatives, whereas redundancy-reduction-based methods \cite{ermolov2021whitening, zbontar2021barlow, bardes2022vicreg, bardes2022vicregl} employ principled regularization terms to minimize features redundancy. For instance, in \cite{zbontar2021barlow} a loss is introduced to minimize the cross-correlation between features of the positive pairs. Similarly, in \cite{bardes2022vicreg, bardes2022vicregl} the features learning process minimizes the covariance alongside regulating the variance of the embeddings.

Clustering-based methods~\cite{caron2018deep,li2021prototypical,asano2020self,caron2020unsupervised,caron2021emerging}, instead, naturally discretize the latent space via clustering. They perform clustering either offline (using the whole dataset), as in DeepCluster v2~\cite{caron2018deep}, PCL~\cite{li2021prototypical} and SeLa~\cite{asano2020self}, or online (using mini-batches), as in SwAV~\cite{caron2020unsupervised} and DINO~\cite{caron2021emerging}, and impose coherent clustering assignments of positive pairs through a cross-entropy. Noticeably, this loss contrasts targets (assignments) and predictions as in the standard supervised learning objective. Nonetheless, surprisingly, to the best of our knowledge there have been no previous attempts in the literature to exploit this favorable property in the semi-supervised scenario. Our work aims to fill this gap.

\paragraph{Semi-supervised learning.}
Semi-supervised approaches aim to exploit a limited amount of annotations and a large collection of unlabeled data. The most intuitive approach to this task is perhaps Pseudo-Labels~\cite{lee2013pseudo}, based on self-training via pseudo-labeling: a model trained on labeled data generates categorical pseudo-labels for the unlabeled examples, which will be then integrated into the labeled set for the next model training. However, hard (categorical) labels easily exacerbate the classification bias of the training model, a phenomenon known as \textit{confirmation bias}~\cite{arazo2020pseudo}. To counteract this issue, researchers have shown benefits from soft labels and confidence thresholding \cite{arazo2020pseudo} as well as from different training strategies like co- and tri-training~\cite{qiao2018deep, nassar2021all, dong2018tri}, model distillation~\cite{xie2020self}, consistency regularization (see below) and model de-biasing \cite{schmutz2022don, wang2022debiased}.

Consistency regularization methods operate by introducing additional losses computed on unsupervised samples and enforce consistency of the network output under perturbation of the model and/or the input~\cite{sajjadi2016regularization, laine2017temporal, miyato2016distributional, park2018adversarial, tarvainen2017mean, athiwaratkun2019there, zhang2020wcp, verma2019interpolation,xie2020unsupervised}. Recent approaches integrate pseudo-labeling techniques and consistency regularization. FixMatch~\cite{sohn2020fixmatch} generates pseudo-labels from weak perturbations of the input that are used as target for strong input perturbations whenever they satisfy an arbitrary confidence threshold. \cite{berthelot2019mixmatch, berthelot2020remixmatch, liu2022decoupled} exploit MixUp~\cite{zhang2018mixup} to improve class boundaries in low-density regions. Other works explore more advanced pseudo-labeling techniques based on adaptive confidence thresholds~\cite{xu2021dash, zhang2021flexmatch} and meta-learning~\cite{pham2021meta}, uncertainty estimation~\cite{chen2020semi, wang2022np}, latent structure regularization~\cite{li2021comatch}. Recently, ConMatch~\cite{kim2022conmatch} extended prediction consistency with self-supervised features consistency, while SimMatch~\cite{zheng2022simmatch} improved consistency regularization by applying it at both semantic-level and instance-level. Class-aware Contrastive Semi-Supervised Learning (CCSSL) is introduced in \cite{yang2022class} to improve the quality of pseudo-labels in presence of unknown (out-of-distribution) classes.

As self-supervised methods have the ability to extract relevant information from unlabeled data, they can be leveraged to tackle the semi-supervised problem. Noticeably, self-supervised pre-training is found beneficial for many consistency regularization methods~\cite{cai2021exponential}. Moreover, semi-supervised self-supervised methods exist, which consider multi-tasking~\cite{zhai2019s4l, wang2020enaet}, pre-training distillation~\cite{chen2020big}, exponential moving average normalization~\cite{cai2021exponential} and supervised contrastive~\cite{assran2020supervision}. Finally, PAWS~\cite{assran2021semi} borrows some principles from self-supervised clustering, but, somewhat similarly to SimMatch~\cite{zheng2022simmatch}, it assumes labeled instances as anchors to compare views. However, it does not exploit the multitasking of the self- and semi- supervised objectives nor takes advantage of latent clustering.\begin{figure*}[t]
    \minipage{.63\textwidth}
    \centering
    \vspace{6mm}
        \includegraphics[width=\linewidth]{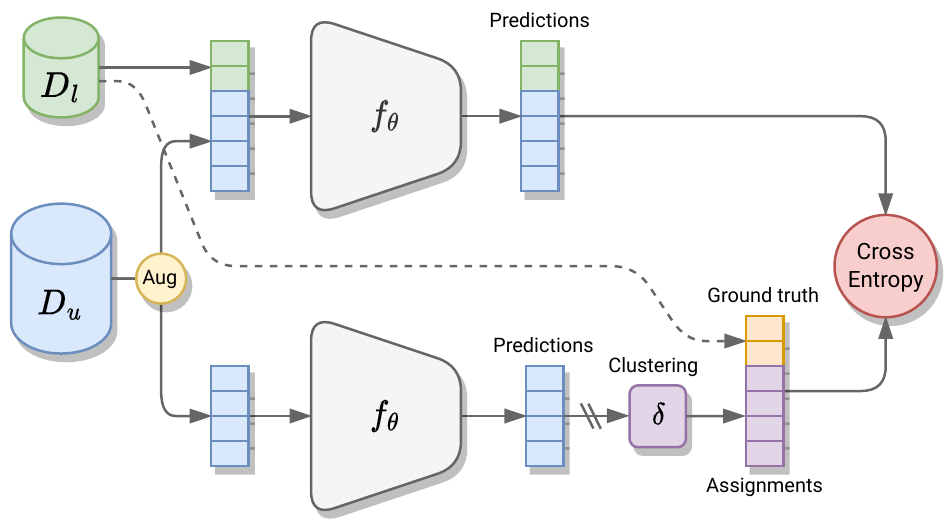}
    \endminipage
    \,
    \minipage{.36\textwidth}
    \centering
        {\input{figures/algo.tex}}
    \endminipage
    \vspace{-2mm}
    \caption{Left: overview of the proposed family of semi-supervised methods. Right: PyTorch-like pseudo-code for Suave and Daino. }
    \label{fig:method}
    \vspace{-2mm}
\end{figure*}

\section{Clustering-based semi-supervised learning}
\label{sec:method}
Clustering-based self-supervision is typically adopted in unsupervised representation learning scenarios where no labeled data is available \cite{caron2018deep,caron2020unsupervised,caron2021emerging}. However, in this work, we aim to take advantage of the few annotations available in the semi-supervised setting to learn even better representations. Our main intuition is to replace the cluster centroids with class prototypes learned with supervision. In this way, unlabeled samples will be clustered around the class prototypes, guided by the self-supervised clustering-based objective. To this end, we jointly optimize a supervised loss on the labeled data and a self-supervised loss on the unlabeled data. It turns out that using the same loss function (cross-entropy) is a very good choice since it promotes synergy between the two objectives and eases up the implementation. In the following, we formalize self-supervised clustering in Section~\ref{ssec:deep_cluster}. Then, we describe the details of our novel semi-supervised learning framework in Section~\ref{ssec:semisup_cluster} and show its application through two popular self-supervised approaches in Section~\ref{ssec:suavendino}. %

\subsection{Clustering-based self-supervised learning}
\label{ssec:deep_cluster}

Given an unlabeled dataset $\train_u = \{\rvx_u^1, ..., \rvx_u^N\}$, two correlated views of the same input image, $(\rvx, \hat{\rvx})$, are generated via data augmentation and embedded through an encoder network $f_{\theta}$, composed of a backbone $g$, a projector $h$ and a set of prototypes (or centroids) $p$ implemented as a bias-free linear layer. Performing the forward pass of the backbone and the projector produces two latent representations, $(\rvh, \hat{\rvh})$ for the two correlated views $(\rvx, \hat{\rvx})$, respectively. Subsequently, the cluster centroids $p$ are used to produce two sets of logits $(\rvp, \hat{\rvp})$ corresponding to each of the two latent representations. A softmax function $\sigma$ can be applied to these logits to obtain probability distributions over the set of clusters, also referred to as cluster assignments. 

In principle, the two assignments could be immediately compared in order to encourage the network to output similar predictions for similar inputs (correlated views). However, this may lead to degenerate solutions where all samples are assigned to the same cluster. To avoid collapse, simple clustering techniques are usually employed to embed priors into the cluster assignment process and regularize the training procedure. In practice, we compute:
\begin{equation}
\label{eq:delta}
\hat{\rvy} = \delta(\hat{\rvp}, \rvc, \epsilon),
\end{equation}
where $\delta$ is the clustering technique, which takes as input some predicted logits $\hat{\rvp}$, a context $\rvc$ and a temperature-like parameter $\epsilon$ that usually regulates the entropy of the assignment $\hat{\rvy}$ (sometimes addressed as pseudo-label). The context $\rvc$ can be implemented in different flavors depending on the nature of $\delta$. For instance, for offline clustering $\rvc$ is represented by the features of all the samples in the dataset, while for online clustering the context might contain the features of the current batch or simply a running mean of the overall distribution. The pseudo-label can be categorical (hard) or soft. Empirically soft cluster assignments were shown to yield superior performance~\cite{caron2020unsupervised}. 

The output of the clustering technique is then used as a target in the cross-entropy loss:
\begin{equation}
\label{eq:loss_self}
\ell(\rvs,  \hat{\rvy})=-\sum_{k=1}^K \hat{\rvy}_k \log \left(\rvs_k\right),
\end{equation}
where $\rvs = \sigma(\rvp / \tau)$ has gone through softmax normalization with temperature $\tau$, and $K$ is the number of clusters. Note that a stop-gradient operation is performed, so that the gradient is not propagated through the pseudo-label. It is worth mentioning that the cross-entropy loss in Eq.~\ref{eq:loss_self} is asymmetric, but, it can be made symmetric by swapping $\rvx$ with $\hat{\rvx}$.

Intuitively, this loss leverages cluster assignments as a proxy to minimize the distance between latent representations $(\rvh, \hat{\rvh})$ of augmented views of the same image. As a by-product, the objective also learns a set of cluster prototypes encoded in the last linear layer $p$, which represent semantic information in the latent space. However, these clusters have no guarantee to be aligned with the true semantic categories represented in the dataset. Nonetheless, the discretization of the feature space is particularly interesting in the context of semi-supervised learning, as described in the following.

\subsection{Our semi-supervised learning framework}
\label{ssec:semisup_cluster}
In the semi-supervised scenario, we assume having access to a partially labeled dataset, $\train = \train_u \cup \train_l$, usually with $|\train_l| << |\train_u|$ and $\train_l$ contains a number of $C$ known classes. We propose to exploit clustering-based self-supervised models described in  Sec.~\ref{ssec:deep_cluster} as a base to extract information from unlabeled data, and we extend them to take advantage of the labeled samples.

As mentioned, the main drawback of clustering-based self-supervised methods in the context of semi-supervised learning is that there are no guarantees that the stochastic optimization process will organize the clusters in the feature space according to the class labels. 
Indeed, they may be completely misaligned with the actual distribution of the classes. This also potentially hinders the effectiveness of the representations, as some prototypes may encode spurious correlations in the data. %
An ideal scenario is one where the clusters are centered on the actual class centroids such that the label can be propagated to the unlabeled samples by means of the clustering function. This will generate positive feedback that progressively transfers information from the labeled set into the unlabeled one, thus improving the feature representations learned by the network.

We propose to condition the cluster prototypes to encode the class information by resorting to multi-tasking of the self-supervised and supervised objectives. This can be achieved by optimizing the same loss function in Eq.~\ref{eq:loss_self} while replacing the pseudo-label with the ground-truth label when available:
\begin{equation}
\label{eq:loss_semi}
\ell(\rvs,  \bar{\rvy}),\quad \boldsymbol{\bar{\rvy}}= \begin{cases}\rvy, & \rvx \in \train_l \\ \hat{\rvy}, & \rvx \in \train_u.\end{cases}
\end{equation}
Since the linear layer $p$ that contains the prototypes is now shared between the two objectives, we set $K=C$ to have a matching label space, although in principle this is not a hard constraint (as shown in \cite{fini2021unified}). In a nutshell, in our framework we compute the forward pass for both labeled and unlabeled samples, concatenate the associated predictions and the targets and apply the cross-entropy loss simultaneously as described in Fig.~\ref{fig:method}. We empirically demonstrate in Sec.~\ref{sec:exp} that, despite its simplicity, our framework equipped with these design choices is a strong semi-supervised learner. 

\subsection{Suave and Daino}
\label{ssec:suavendino}
Our proposed framework described above can convert any clustering-based self-supervised method into a semi-supervised learner, without dropping, adding or replacing any architectural components and reusing the same loss function. We select two representative self-supervised methods to showcase our framework: SwAV~\cite{caron2020unsupervised} and DINO~\cite{caron2021emerging}, whose semi-supervised extensions we name \textbf{Suave} and \textbf{Daino}. This choice is motivated by their superior representation learning capabilities and ease of use. In particular, both are online clustering methods, which means that the cluster assignments can be computed on-the-fly without accessing the whole dataset simultaneously. This is a great advantage, especially for large scale datasets. For these reasons, we discard offline clustering methods like DeepCluster~\cite{caron2018deep} and PCL~\cite{li2021prototypical}, while in principle our approach can be also applied to them. 

\noindent \textbf{Suave.} SwAV~\cite{caron2020unsupervised}, following~\cite{asano2020self}, casts the pseudo-label assignment problem as an instance of the optimal transport problem and proposes the swapped prediction task where the assignment of a view is predicted from the representation of another view. Simply put, SwAV generates pseudo-labels such that each cluster is approximately equally represented in the current batch, preventing the network from falling into degenerate solutions. This is especially convenient as we only need the information in the current batch for clustering. In light of our proposed framework, reusing Eq.~\ref{eq:delta}, in Suave the target for the first sample in the batch can be obtained as follows (the same reasoning can be trivially applied to the other samples in the batch):
\begin{equation}
    \hat{\rvy}_1 = \delta(\hat{\rvp}_1, [\hat{\rvp}_2, ..., \hat{\rvp}_{B}], \epsilon)
\end{equation}
where the context $\rvc = [\hat{\rvp}_2, ..., \hat{\rvp}_{B}]$ contains all the logits in the batch except $\hat{\rvp}_1$. Now we define $\hat{\bm{P}} = [\hat{\rvp}_1, \hat{\rvp}_2, \dots, \hat{\rvp}_B]$ and $\hat{\bm{Y}} = [\hat{\rvy}_1, \hat{\rvy}_2, \dots, \hat{\rvy}_B]^\top$, where $\hat{\bm{Y}}$ is the matrix that holds the unknown pseudo-labels of the whole batch. The clustering function $\delta$ will return the first column of $\hat{\bm{Y}}$ that is found by solving:
\begin{equation}
\label{eq.Swav-optimization} 
\hat{\bm{Y}} = \max_{\bm{Y} \in \Gamma}  \Tr (\bm{Y} \hat{\bm{P}}) + \epsilon \operatorname{H}(\bm{Y}),
\end{equation}
\noindent
where $\epsilon\! >\! 0$ is the temperature-like parameter mentioned in Eq.~\ref{eq:delta}, $\operatorname{H}$ is the entropy function, $\Tr$ is the trace operator, and $\Gamma$ is the transportation polytope defined as:
\begin{equation}
\small
\label{eq.polytope} 
 \Gamma = \{ \bm{Y} \in \mathbb{R}^{C \times B}_+ | \bm{Y}  \bm{1}_B = \frac{1}{C} \bm{1}_{C}, 
 \bm{Y}^\top  \bm{1}_{C} = \frac{1}{B} \bm{1}_{B} \}.
\end{equation}
These constraints enforce that, on average, each cluster is selected $\frac{B}{C}$ times in each batch, automatically ensuring de-biased assignments. The solution to Eq.~\ref{eq.Swav-optimization} is obtained using the Sinkhorn-Knopp algorithm \cite{cuturi2013sinkhorn,amrani2022self}.

\noindent \textbf{Daino.} DINO~\cite{caron2021emerging} aims at further simplifying the pipeline described above. Instead of using optimal transport on the predictions of the current batch it uses two practical tricks to avoid collapse: a momentum encoder and a pseudo-labeling strategy based on centering and sharpening. A momentum encoder is a ``slow'' version of the encoder $f_{\theta}$ updated using exponential moving average (EMA). After each gradient step on $\theta$, the parameters $\phi$ of the momentum encoder $f_{\phi}$ are updated as follows:
\begin{equation}
    \phi \leftarrow \eta \phi + (1-\eta) \theta,
\end{equation}
where $\eta$ is a rate parameter. The rationale behind this choice is that the momentum encoder serves as a teacher producing more stable representations throughout training, improving the optimization process. The teacher is used at every iteration to generate the logits $\hat{\rvp} = f_\phi(\hat{\rvx})$ which in turn are input to the clustering function $\delta$ in Eq.~\ref{eq:delta} to obtain the target assignment for Daino:
\begin{equation}
    \hat{\rvy} = \delta(\hat{\rvp}, \gamma, \epsilon) = \sigma\left(\frac{\hat{\rvp} - \gamma}{\epsilon}\right),
\end{equation}
where $\sigma$ and $\epsilon$ are the softmax function and a temperature coefficient (here used to sharpen the distribution) respectively. The context $\rvc = \gamma$ in this case is a centering vector that approximates and de-biases the overall distribution of the data over the clusters and is also updated using EMA:
\begin{equation}
    \gamma \leftarrow \mu \gamma + (1-\mu) \frac{1}{B} \sum_{b=0}^{B}\hat{\rvp}_b,
\end{equation}
where $\mu$ adjusts the rate of the update. In brief, centering prevents one dimension to dominate but encourages high-entropy outputs, while sharpening does the opposite. Empirical evidence shows that this is enough to avoid collapse.

\section{Implementation details}
\label{sec:implementation}

\paragraph{Architectures.}
For large-scale datasets (ImageNet), we adopt  ResNet50~\cite{he2016deep} and ViT-S/16~\cite{dosovitskiy2021image} backbones. For small-scale datasets (CIFAR100) we train a Wide ResNet (WRN-28-8)~\cite{zagoruyko2016wide}. The convolutional backbone is followed by a projection head consisting of a multi-layer perceptron (with batch normalization in the hidden layers). We use 2 layers as in \cite{caron2020unsupervised,assran2021semi}. After that, we perform L2-normalization and compute the predictions using a bias-free L2-normalized linear layer corresponding to the prototypes. We set the number of prototypes equal to the number of classes of the dataset at hand. Moreover, we perform an online linear evaluation at different depths of the network to identify which layer learns the best representations; we append a detached classification head on top of the backbone, the first and the second layer of the projection. We remark that such heads do not impact the efficiency of the training. 

\noindent \textbf{Semi-supervised pre-training.}
We pre-train our models using LARS~\cite{you2017large} optimizer with linear warmup plus cosine learning rate schedule. Also, we adopt weight decay regularization. The backbone layers of the models can be initialized with self-supervised checkpoints of SwAV and DINO. At each training iteration, we sample a mini-batch composed of unlabeled images\footnote{We sample unlabeled examples from dataset $\train$, instead of $\train_u$.} and labeled images. Note that we count a training epoch considering a full pass over the unlabeled dataset. We optimize the cross-entropy loss of Eq.~\ref{eq:loss_semi}, re-weighting the labeled and unlabeled terms by their frequency in the batch. Moreover, we soften the supervised targets with label smoothing to mitigate overfitting. For the pseudo-labeling, in Suave, we re-use the same parameters for the Sinkhorn-Knopp as in SwAV; in Daino, we inherit the hyperparameters for the centering and the momentum encoder, whereas we tune the sharpening coefficient. More details are in the supplementary material.

\noindent \textbf{Data augmentation.}
Images are augmented differently based on whether they are unlabeled or labeled. For the unlabeled images, we follow the default self-supervised augmentations of SwAV and DINO, while for the labeled, we adopt lighter augmentation Inception-style (random crop and flip)~\cite{szegedy2015going} and color distortion (jittering and greyscale). Note that it is important not to over-distort the labeled images as they are needed to align the clusters and classes.

To boost the self-supervised feature learning, we employ the multi-crop~\cite{caron2020unsupervised} augmentation scheme for unlabeled images. From each input image, we derive two global views from larger and higher-resolution crops and multiple smaller views from tighter crops. This is common practice in self-supervised learning. Similar to SwAV and DINO, we compute a clustering assignment only for the global views and use them as targets for the smaller ones. All the multi-crop views are taken into account when weighting the loss.

Another augmentation technique we empirically found to be useful is based on the combination of CutMix~\cite{yun2019cutmix} and MixUp~\cite{zhang2018mixup}. We apply it to both unlabeled (global views only) and labeled images, but separately. For the unlabeled images, we interpolate the learned clustering assignments due to the lack of ground-truth labels. This augmentation allows for shifting decision boundaries to low-density regions of the data ~\cite{berthelot2019mixmatch, verma2019interpolation}. We mix the whole batch at every iteration, but to not over-regularize, we concatenate the mixed images to the current batch, instead of substituting it.

\noindent \textbf{Semi-supervised fine-tuning.}
Since self-supervised methods are trained with strong augmentations, it is preferable after pre-training to fine-tune them to slightly improve the performance. We discover that a semi-supervised fine-tuning recipe works better than the typical fully supervised fine-tuning \eg \cite{assran2020supervision}. In practice, we just keep training the model with the
same objective (our semi-supervised clustering-based loss)
for a few more epochs, while relaxing some of the stronger augmentations adopted during pre-training, \ie disable multi-crop and color distorsions.

\section{Experiments}
\label{sec:exp}

\subsection{Experimental protocol}
\label{ssec:exp_proto}

\noindent \textbf{Datasets.}
We perform our experiments using two common datasets, {i.e.}, CIFAR100~\cite{krizhevsky2009learning} and ImageNet-1k~\cite{deng2009imagenet}. In the semi-supervised setting the training set of each of these datasets is split into two subsets, one labeled and one unlabeled. For CIFAR100, which is composed of 50K images equally distributed into 100 classes, we investigate three splits as in~\cite{sohn2020fixmatch}, retaining 4 (0.8\%), 25 (5\%), and 100 (20\%) labels per class, resulting in total to 400, 2500, and 10000 labeled images, respectively. For ImageNet-1k ($\sim$1.3K images per class, 1K classes), we adopt the same two splits of \cite{chen2020big} using 1\% and 10\% of the labels. In both datasets, we evaluate the performance of our method by computing top-1 accuracy on the respective validation/test sets. 

\noindent \textbf{Baselines.}
We compare our methods, Suave and Daino, with state of the art methods from the semi-supervised literature (see Section~\ref{sec:rel_works}). In particular, we compare against hybrid consistency regularization methods like SimMatch~\cite{zheng2022simmatch} and ConMatch~\cite{kim2022conmatch} (and others~\cite{zhang2021flexmatch, wang2022np, liu2022decoupled}), and methods that are derived from self-supervised approaches, like PAWS~\cite{assran2021semi} and S$^4$L-Rot~\cite{zhai2019s4l}. We also compare with recent debiasing-based pseudo-labeling methods like DebiasPL~\cite{wang2022debiased}. For the sake of fairness, we leave out methods using larger architectures or pre-trained on larger datasets, \eg DebiasPL with CLIP~\cite{radford2021learning} and SimCLR v2~\cite{chen2020big}.

\subsection{Results}
\label{ssec:res}
First, we demonstrate the effectiveness of our semi-supervised framework in a small-scale dataset using the CIFAR100 benchmark. Then, we evaluate our models at large-scale on ImageNet-1k (see Section~\ref{ssec:res_cifar}). Finally, we ablate the different components of our models (see Section~\ref{ssec:ablations}). 
\subsubsection{Comparison with the state of the art}
\label{ssec:res_cifar}
\begin{table}[t]
\centering
\caption{Comparison with the state-of-the-art on CIFAR100.}
\vspace{-2mm}
\label{tab:cifar100}
\begin{tabular}{lccc}
\toprule
\multirow{2}[1]{*}{Method} & \multicolumn{3}{c}{\textbf{Acc@1}}  \\ 
\cmidrule(r){2-4}
                        & 400           & 2500          & 10000         \\ \midrule
$\Pi$-Model\cite{laine2017temporal}               &  -         & 42.8          & 62.1          \\
Mean Teacher~\cite{tarvainen2017mean}               &  -         & 46.1          & 64.2          \\
MixMatch~\cite{berthelot2019mixmatch}               &  32.4         & 60.2          & 72.2          \\
UDA~\cite{xie2020unsupervised}               &  53.6         & 72.3          & 77.5          \\
ReMixMatch~\cite{berthelot2020remixmatch}               &  55.7         & 72.6          & 77.0          \\
FixMatch~\cite{sohn2020fixmatch}               &  50.1         & 71.4          & 76.8          \\
Dash~\cite{xu2021dash}               & 55.2          & 72.8          & 78.0          \\
CoMatch~\cite{li2021comatch}             & 60.0          & 73.0          & 78.2          \\
Meta Pseudo Labels~\cite{pham2021meta}             & 55.8          & 72.3          & 77.5          \\
FlexMatch~\cite{zhang2021flexmatch}               & 60.1          & 73.5          & 78.1          \\
FixMatch+DM~\cite{liu2022decoupled}             & 59.8          & 74.1          & 79.6          \\
NP-Match~\cite{wang2022np}                & 61.1          & 74.0          & 78.8          \\
ConMatch~\cite{kim2022conmatch}                & 61.1          & 74.6          & -             \\
SimMatch~\cite{zheng2022simmatch}                & 62.2          & 74.9          & 79.4          \\
CCSSL~\cite{yang2022class}                & 61.2          & 75.7          & 80.1             \\
\rowcolor{cyan!20}Daino           & 61.1          & 75.2          & 79.2          \\
\rowcolor{cyan!20}Suave          & \textbf{64.6} & \textbf{77.0} & \textbf{81.6} \\ \bottomrule
\end{tabular}%
\vspace{-2mm}
\end{table}
\noindent \textbf{Results on CIFAR100.} Table~\ref{tab:cifar100} shows a comparison between our methods and several semi-supervised approaches in the literature. In particular, we compare to consistency regularization semi-supervised methods (see Section~\ref{sec:rel_works}) like ConMatch~\cite{kim2022conmatch} and SimMatch~\cite{zheng2022simmatch}, which are the strongest methods on this dataset.
First, from the table we observe that both Suave and Daino achieve high performance in all the three splits (400, 2500, 10000). Daino obtains results comparable to the best competitors, while Suave outperforms all the baselines and beats the best methods CCSSL~\cite{yang2022class} and SimMatch~\cite{zheng2022simmatch} by +2.4\%p, +1.3\%p, +1.5\%p in the three settings, respectively. 
Overall, these results clearly demonstrate how our clustering-based semi-supervised learning methods achieves state-of-the-art performance without requiring any \textit{ad-hoc} confidence thresholds for pseudo-labels as in most recent consistency regularization methods. 
Interestingly, comparing Suave and Daino with their self-supervised counterparts, SwAV and DINO, a remarkable improvement is achieved. In fact, SwAV and DINO obtain an accuracy of 64.9\% and 66.8\%~\cite{da2022solo}, respectively, when linearly evaluated using 100\% of the labels after self-supervised pre-training.

\begin{table}[t]
\centering
\caption{Comparison with the state-of-the-art on ImageNet-1k with ViT-S/16~\cite{dosovitskiy2021image}. DINO and MSN perform linear evaluation with labeled data on top of frozen features.}
\vspace{-2mm}
\label{tab:in_vit_results}
\resizebox{\linewidth}{!}{%
\begin{tabular}{lccccc}
\toprule
\multirow{2}[1]{*}{\textbf{Method}} & \multirow{2}[1]{*}{\textbf{Epochs}} & \multicolumn{2}{c}{\textbf{Batch size}} & \multicolumn{2}{c}{\textbf{Acc@1}} \\
\cmidrule(r){3-4}\cmidrule(r){5-6}
&  & Unlab. & Lab.  & 10\% & 1\%  \\
\midrule
DINO~\cite{caron2021emerging} & (800) & 1024 & - & 72.2 & 64.5 \\
MSN~\cite{assran2022masked} & (800) & 1024 & - & - & \textbf{67.2}  \\
\rowcolor{cyan!20} Daino & (800) 60 & 1024 & 512 & \textbf{76.6} & 67.1 \\
\bottomrule
\end{tabular}%
}
\vspace{-4mm}
\end{table}

\begin{table}[t]
\centering
\caption{Comparison with the state-of-the-art on ImageNet-1k. All the models reported use ResNet-50. In the first and the second column, we indicate within brackets whether a model is initialized from a self-supervised checkpoint and the number of epochs of that pre-training. For brevity, we refer to FixMatch as FM. $^*$SimMatch does not report the batch size, and its value is inferred from the public repository. $\dagger$ refers to epochs on labeled data.}
\vspace{-2mm}
\label{tab:in_results}
\resizebox{\linewidth}{!}{%
\begin{tabular}{lccccc}
\toprule
\multirow{2}[1]{*}{\textbf{Method}} & \multirow{2}[1]{*}{\textbf{Epochs}} & \multicolumn{2}{c}{\textbf{Batch size}} & \multicolumn{2}{c}{\textbf{Acc@1}} \\
\cmidrule(r){3-4}\cmidrule(r){5-6}
&  & Unlab. & Lab.  & 10\% & 1\%  \\
\midrule
\multicolumn{6}{c}{\textit{with similar batch size and number of epochs}} \\
\midrule
S$^4$L-Rotation~\cite{zhai2019s4l} & 200$^\dagger$ & 256 & 256 & 61.4 & - \\ 
FM-DA ({\footnotesize MoCo v2})~\cite{li2021comatch} & (800) 400 & 640  & 160 & 72.2 & 59.9 \\ 
PAWS~\cite{assran2021semi} & 100 & 256  & 1680 & 70.2 & - \\
CoMatch ({\footnotesize MoCo v2})~\cite{li2021comatch} & (800) 400 & 640  & 160 & 73.7 & 67.1 \\ 
FM-EMAN ({\footnotesize MoCo-EMAN})~\cite{cai2021exponential} & (800) 300 & 320 & 64 & 74.0 & 63.0 \\
SimMatch~\cite{zheng2022simmatch} & 400 & 320$^*$ & 64$^*$ & 74.4 & \textbf{67.2} \\
\multirow[t]{3}{*}{DebiasPL ({\footnotesize MoCo-EMAN})~\cite{wang2022debiased}} & (800) 50 & 640 & 128 & - & 65.3 \\
 & (800) 200 & 640 & 128 & - & 66.5 \\
\rowcolor{cyan!20} \multirow[t]{3}{*}{Suave}    & (100) 100   & 256  & 128  & 73.6 & 63.8 \\
\rowcolor{cyan!20}    & (200) 100   & 256   & 128  & 74.3 & 65.0 \\
\rowcolor{cyan!20}    & (800) 100   & 256   & 128  & \textbf{75.0} & 66.2 \\ 
\midrule
\multicolumn{6}{c}{\textit{with larger batch size or number of epochs}} \\
\midrule
UDA~\cite{sohn2020fixmatch} & $\sim$480 & 15360  & 512 & 68.8 & - \\ 
Meta Pseudo Labels~\cite{pham2021meta} & $\sim$800 & 2048  & 2048 & 73.9 & - \\ 
\multirow[t]{3}{*}{PAWS\cite{assran2021semi}}
 & 100  & 4096  & 6720 & 73.9 & 63.8 \\
 & 200  & 4096  & 6720 & 75.0 & 66.1 \\
 & 300  & 4096  & 6720 & 75.5 & 66.5 \\
DebiasPL ({\footnotesize MoCo-EMAN})~\cite{wang2022debiased} & (800) 300 & 1280 & 256 & - & 67.1 \\
\midrule
\multicolumn{6}{c}{\textit{self-supervised pre-training with fine-tuning}} \\
\midrule
MoCo v2~\cite{chen2020improved} & (800) & 256  & - & 66.1 & 49.8 \\
SimCLR v2~\cite{chen2020simple} & (1000) & 4096  & - & 68.4 & 57.9 \\
BYOL~\cite{grill2020bootstrap} & (1000) & 2048  & - & 68.8 & 53.2 \\
MoCo-EMAN~\cite{cai2021exponential} & (800) & 256  & - & 68.1 & 57.4 \\
SwAV~\cite{cai2021exponential} & (800) & 4096  & - & 70.2 & 53.9 \\
NNCLR~\cite{dwibedi2021little} & (1000) & 4096  & - & 70.2 & 56.4 \\
Barlow Twins~\cite{zbontar2021barlow} & (1000) & 2048  & - & 69.7 & 55.0 \\
FNC~\cite{huynh2022boosting}  & (1000) & 4096  & - & 71.1 & 63.7 \\
\bottomrule
\end{tabular}%
}
\vspace{-5mm}
\end{table}
\noindent \textbf{Results on ImageNet-1k.} We also perform large scale experiments on ImageNet-1k considering the label splits of 1\% and  10\% as common in previous works \cite{zheng2022simmatch,sohn2020fixmatch}. Due to limited computational resources, we primarily focus on Suave with a ResNet50 backbone, which enables us to compare with most of the related state-of-the-art methods. In addition, we also provide experimental evidence that our framework works well with a different  clustering algorithm (Daino), backbone (ViT~\cite{dosovitskiy2021image}), and a simpler training recipe (described in the supplementary meterial).

\begin{figure}
    \centering
    \includegraphics[width=.48\columnwidth]{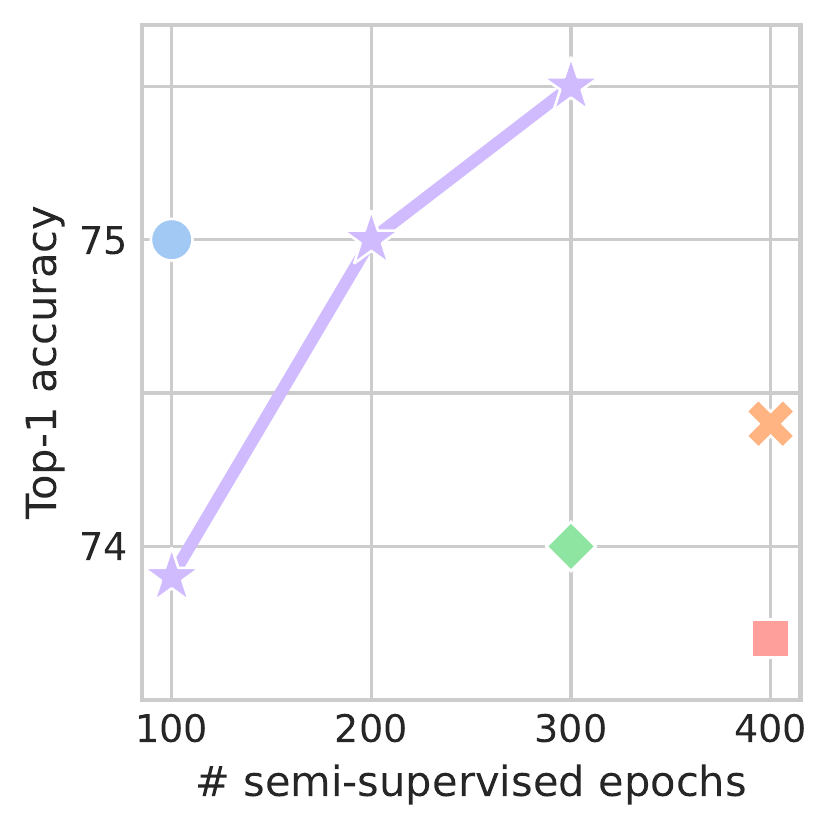}
    \hfill
    \includegraphics[width=.49\columnwidth]{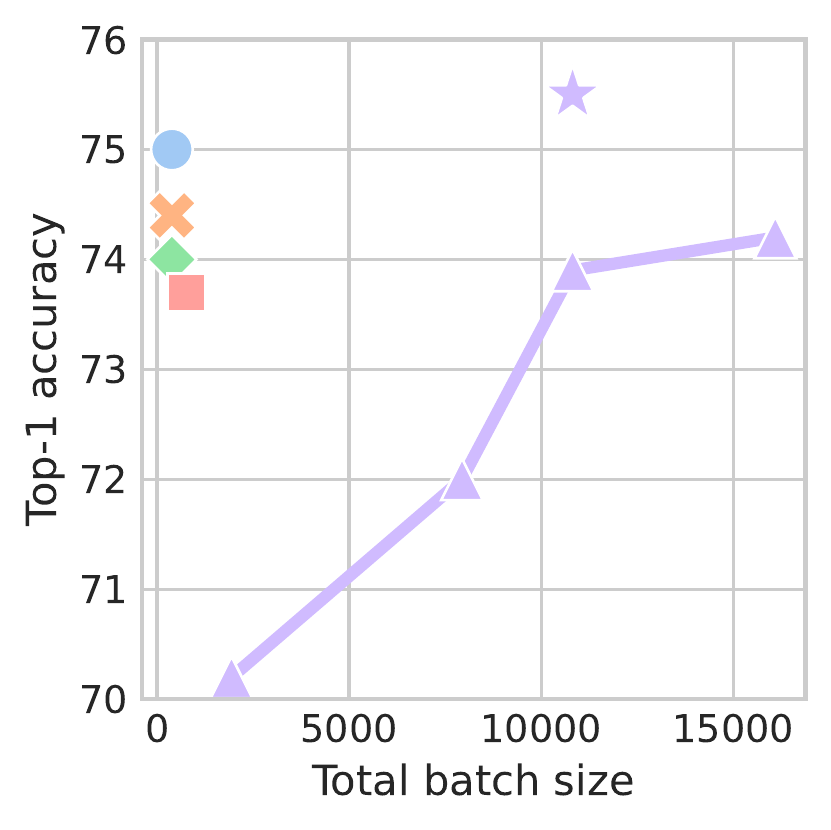}
    \vspace{-2.5mm}
    \includegraphics[width=.8\columnwidth]{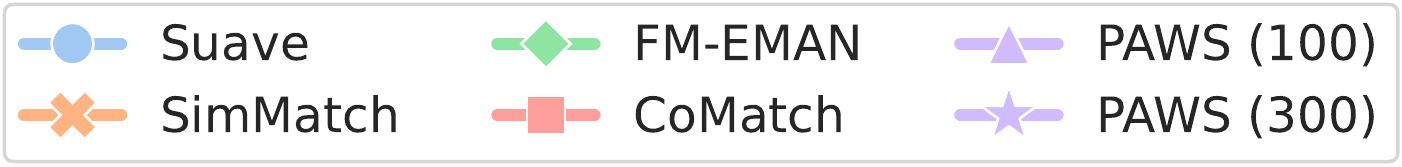}
    \caption{Comparison with related works on the training efficiency in terms of pre-training epochs (top) and size of the mini-batches (bottom). Plotted results refer to the ImageNet 10\% split.}
    \vspace{-3.5mm}
    \label{fig:acc_vs_bs}
\end{figure}

We compare against three families of methods, self-supervised-inspired semi-supervised approaches \cite{zhai2019s4l, assran2021semi}, consistency regularization methods \cite{zheng2022simmatch,sohn2020fixmatch,li2021comatch, cai2021exponential, xie2020unsupervised, pham2021meta}, and debiasing-based methods \cite{wang2022debiased}, and present results in Table~\ref{tab:in_results}, Table~\ref{tab:in_vit_results}, and Figure~\ref{fig:acc_vs_bs}. To provide full context to our results, in Table~\ref{tab:in_results}, we also report the performance of self-supervised models when simply fine-tuned with labels.

Suave and Daino obtain performances that are comparable with state-of-the-art methods on ImageNet-1k. In particular, when compared against related methods (\eg DebiasPL, SimMatch, FixMatch-EMAN, CoMatch) with similar batch-size and number of epochs, Suave obtains the best score (+0.6\%p) on the 10\% setting and the third best score on the 1\% setting, -1.0\%p from the best baseline SimMatch~\cite{zheng2022simmatch}. Importantly, other approaches like PAWS~\cite{assran2020supervision} require larger batches to obtain comparable results. This aspect is even more evident by looking at Fig.~\ref{fig:acc_vs_bs} (bottom), where PAWS significantly underperforms with respect to our method when decreasing the batch size. This behavior can be ascribed to the fact that PAWS adopts a k-nearest neighbor approach on the labeled instances to generate the assignment vectors (pseudo-labels), and thus it requires a sufficiently high number of labeled examples to well represent the class distributions. In contrast, Suave generates the assignments using the learnable cluster/class prototypes, which are a fixed number independent of the batch size.

Figure~\ref{fig:acc_vs_bs} (top) compares Suave with the best methods in Table~\ref{tab:in_results}, highlighting the fast convergence of our method. Performing 100 epochs of semi-supervised pre-training is enough to achieve comparable results to other related methods like FixMatch-EMAN, PAWS, and SimMatch, which require at least twice the semi-supervised epochs. It is worth noting that Suave naturally benefits from self-supervised initialization, as it basically shares the objective with the self-supervised counterpart 
SwAV. 
Indeed, by looking at Table~\ref{tab:in_results} we can observe that better SwAV checkpoints lead to higher accuracy. At the same time, the difference between Suave (SwAV-100) and Suave (SwAV-800) is rather small, 1.4\%p and 2.4\%p on 10\% and 1\%, respectively.%
\vspace{-1.5mm}
\subsubsection{Ablation study}
\vspace{-1mm}
\label{ssec:ablations}
Here, we assess the importance of the main components of our method: (i) the multi-task training objective, (ii) the adopted fine-tuning strategy, and (iii) the quality of representations at different depth of the network.   

In Table~\ref{tab:abl_loss}, we demonstrate the importance of our multi-tasking strategy. Suave significantly outperforms vanilla SwAV, even in its improved version, \ie SwAV (repro) obtained with the fine-tuning protocol borrowed from  PAWS~\cite{assran2021semi}. More importantly, when comparing Suave to SwAV+CT (SwAV where the self-supervised loss is multi-tasked with a supervised contrastive loss~\cite{khosla2020supervised}) we still observe that our method is far superior.
This clearly indicates that it is better to exploit the available labels to condition the prototypes rather than using them to sample positive instances for contrastive training.
\begin{table}[t]
\centering
\caption{Impact of our multi-task training strategy. Legend: \textbf{UPT} = ``unsupervised pre-training", \textbf{MT}  = ``multi-tasking'' and \textbf{SL}  = ``same loss for all training phases''.}
\vspace{-2.5mm}
\label{tab:abl_loss}
\resizebox{\linewidth}{!}{%
\begin{tabular}{lccccccc}
\toprule
\multirow{2}[1]{*}{\textbf{Method}} & \multirow{2}[1]{*}{\textbf{UPT}} & \multirow{2}[1]{*}{\textbf{MT}} & \multirow{2}[1]{*}{\textbf{SL}} & \multirow{2}[1]{*}{\textbf{Epochs}} & \multicolumn{2}{c}{\textbf{Acc@1}} \\
\cmidrule(r){6-7}
&  &  & &  & 10\% & 1\%  \\
\midrule
SwAV    & \checkmark & & \checkmark & 800   & 70.2 & 53.9 \\
\rowcolor{gray!20}\multirow[t]{3}{*}{SwAV (repro)} & \checkmark	 &  & \checkmark & 800 & 72.3 & 57.0 \\
SwAV+CT~\cite{assran2020supervision} & \checkmark & \checkmark & & (400)30 & 70.8 & - \\
\rowcolor{cyan!20}\multirow[t]{3}{*}{Suave} & \checkmark & \checkmark & \checkmark & (800)100   & \textbf{75.0} & \textbf{66.2} \\ \bottomrule
\end{tabular}%
}
\vspace{-1mm}
\end{table}\begin{table}[t]
\centering
\caption{Impact of our semi-supervised fine-tuning strategy.}
\vspace{-2.5mm}
\label{tab:abl_ft}
\resizebox{\linewidth}{!}{%
\begin{tabular}{lcccc}
\toprule
\multirow{2}{*}{\textbf{Method}} & \multirow{2}{*}{\textbf{Labels (\%)}} & \textbf{Semi-supervised} & \textbf{Supervised} & \textbf{Semi-supervised}\\
&  & \textbf{pre-training} & \textbf{fine-tuning} &  \textbf{fine-tuning} \\
\midrule
\rowcolor{cyan!20}   & 1\% & 64.1 & 64.8 & \textbf{66.2} \\
\rowcolor{cyan!20}  \multirow{-2}{*}{Suave}  & 10\% & 73.4 & 74.8 & \textbf{75.0} \\
\bottomrule
\end{tabular}%
}
\vspace{-1mm}
\end{table}\begin{table}[t]
\centering
\caption{Online linear evaluation on the projector layers. We attach a linear layer (preceded by a stop-grad) after each projector layer and train it with labeled data.}
\vspace{-2.5mm}
\label{tab:abl_proj}
\resizebox{\linewidth}{!}{%
\begin{tabular}{lcccc}
\toprule
\multirow{2}{*}{\textbf{Method}} & \multirow{2}{*}{\textbf{Labels (\%)}} & \multirow{2}{*}{\textbf{Backbone}} & \textbf{Projection} & \textbf{Projection}\\
& &  & \textbf{layer 1} &  \textbf{layer 2} \\
\midrule
\rowcolor{cyan!20}   & 1\% & 65.2 & \textbf{66.2} & 65.1 \\
\rowcolor{cyan!20}  \multirow{-2}{*}{Suave}  & 10\% & 74.2 & \textbf{75.0} & 74.8 \\
\bottomrule
\end{tabular}%
}
\vspace{-3mm}
\end{table}
Table~\ref{tab:abl_ft} provides evidence that fine-tuning Suave with our semi-supervised fine-tuning strategy (see Section~\ref{sec:implementation}) is more effective than adopting the classical fully supervised recipe. We believe that feeding the model with unlabeled data during the fine-tuning phase allows the model to better fit the real distribution. As shown in the table, the semi-supervised fine-tuning recipe enables a larger improvement over the performance after semi-supervised pre-training.

Finally, Table~\ref{tab:abl_proj} shows how the quality of the learned representations changes at different depths of the projector. It turns out that regardless of the label split, we obtain the most discriminative representations at the first layer of the projection. This is consistent with what was observed by other works that use similar augmentations (\eg PAWS~\cite{assran2021semi}). We ascribe this behavior to the fact that the last layer tries to build as much invariance as possible to strong augmentations, which hurts the classification accuracy.
\section{Conclusion}
\vspace{-1mm}
We have presented a novel approach for semi-supervised learning. Our
framework leverages from clustering-based self-supervised methods and adopts a multi-task objective, combining a supervised loss with the unsupervised cross-entropy loss typically adopted for clustering assignments in \cite{caron2020unsupervised,caron2021emerging}.
Despite its simplicity, we demonstrate that our approach is highly effective, setting a new state-of-the-art for semi-supervised learning on CIFAR-100 and ImageNet.
\label{sec:conclusion}\vspace{-3mm}

\scriptsize{\noindent\textbf{Acknowledgements.} This work was supported by the European Institute of Innovation \& Technology (EIT) and the H2020 EU project SPRING (000040103521), funded by the European Commission under GA 871245, and partially supported by the PRIN project LEGO-AI (2020TA3K9N). It was carried out under the ``Vision and Learning joint Laboratory" between FBK and UNITN. Karteek Alahari was funded by the ANR grant AVENUE (ANR-18-CE23-0011). Julien Mairal was funded by the ERC grant number 714381 (SOLARIS project) and by ANR 3IA MIAI@Grenoble Alpes (ANR-19-P3IA-0003). Xavier Alameda-Pineda was funded by the ARN grant ML3RI (ANR-19-CE33-0008-01). This project was granted access to the HPC resources of IDRIS under the allocation 2021-[AD011013084] made by GENCI.}
{\small
\bibliographystyle{ieee_fullname}
\bibliography{egbib}

\begin{thebibliography}{10}\itemsep=-1pt

\bibitem{amrani2022self}
Elad Amrani, Leonid Karlinsky, and Alex Bronstein.
\newblock Self-supervised classification network.
\newblock In {\em European Conference on Computer Vision}, pages 116--132.
  Springer, 2022.

\bibitem{arazo2020pseudo}
Eric Arazo, Diego Ortego, Paul Albert, Noel~E O’Connor, and Kevin McGuinness.
\newblock Pseudo-labeling and confirmation bias in deep semi-supervised
  learning.
\newblock In {\em International Joint Conference on Neural Networks (IJCNN)}.
  IEEE, 2020.

\bibitem{asano2020self}
Yuki~Markus Asano, Christian Rupprecht, and Andrea Vedaldi.
\newblock Self-labelling via simultaneous clustering and representation
  learning.
\newblock In {\em International Conference on Learning Representations}, 2020.

\bibitem{assran2020supervision}
Mahmoud Assran, Nicolas Ballas, Lluis Castrejon, and Michael Rabbat.
\newblock Supervision accelerates pre-training in contrastive semi-supervised
  learning of visual representations.
\newblock {\em arXiv preprint arXiv:2006.10803}, 2020.

\bibitem{assran2022masked}
Mahmoud Assran, Mathilde Caron, Ishan Misra, Piotr Bojanowski, Florian Bordes,
  Pascal Vincent, Armand Joulin, Mike Rabbat, and Nicolas Ballas.
\newblock Masked siamese networks for label-efficient learning.
\newblock In {\em European Conference on Computer Vision}, pages 456--473.
  Springer, 2022.

\bibitem{assran2021semi}
Mahmoud Assran, Mathilde Caron, Ishan Misra, Piotr Bojanowski, Armand Joulin,
  Nicolas Ballas, and Michael Rabbat.
\newblock Semi-supervised learning of visual features by non-parametrically
  predicting view assignments with support samples.
\newblock In {\em Proceedings of the IEEE/CVF International Conference on
  Computer Vision}, 2021.

\bibitem{athiwaratkun2019there}
Ben Athiwaratkun, Marc Finzi, Pavel Izmailov, and Andrew~Gordon Wilson.
\newblock There are many consistent explanations of unlabeled data: Why you
  should average.
\newblock In {\em International Conference on Learning Representations}, 2019.

\bibitem{bachman2019learning}
Philip Bachman, R~Devon Hjelm, and William Buchwalter.
\newblock Learning representations by maximizing mutual information across
  views.
\newblock {\em Advances in neural information processing systems}, 32, 2019.

\bibitem{bardes2022vicreg}
Adrien Bardes, Jean Ponce, and Yann LeCun.
\newblock {VICR}eg: Variance-invariance-covariance regularization for
  self-supervised learning.
\newblock In {\em International Conference on Learning Representations}, 2022.

\bibitem{bardes2022vicregl}
Adrien Bardes, Jean Ponce, and Yann LeCun.
\newblock Vicregl: Self-supervised learning of local visual features.
\newblock In {\em NeurIPS}, 2022.

\bibitem{berthelot2020remixmatch}
David Berthelot, Nicholas Carlini, Ekin~D Cubuk, Alex Kurakin, Kihyuk Sohn, Han
  Zhang, and Colin Raffel.
\newblock Remixmatch: Semi-supervised learning with distribution alignment and
  augmentation anchoring.
\newblock In {\em International Conference on Learning Representations}, 2020.

\bibitem{berthelot2019mixmatch}
David Berthelot, Nicholas Carlini, Ian Goodfellow, Nicolas Papernot, Avital
  Oliver, and Colin~A Raffel.
\newblock Mixmatch: A holistic approach to semi-supervised learning.
\newblock {\em Advances in neural information processing systems}, 32, 2019.

\bibitem{cai2021exponential}
Zhaowei Cai, Avinash Ravichandran, Subhransu Maji, Charless Fowlkes, Zhuowen
  Tu, and Stefano Soatto.
\newblock Exponential moving average normalization for self-supervised and
  semi-supervised learning.
\newblock In {\em Proceedings of the IEEE/CVF Conference on Computer Vision and
  Pattern Recognition}, pages 194--203, 2021.

\bibitem{caron2018deep}
Mathilde Caron, Piotr Bojanowski, Armand Joulin, and Matthijs Douze.
\newblock Deep clustering for unsupervised learning of visual features.
\newblock In {\em Proceedings of the European conference on computer vision
  (ECCV)}, 2018.

\bibitem{caron2020unsupervised}
Mathilde Caron, Ishan Misra, Julien Mairal, Priya Goyal, Piotr Bojanowski, and
  Armand Joulin.
\newblock Unsupervised learning of visual features by contrasting cluster
  assignments.
\newblock {\em Advances in Neural Information Processing Systems},
  33:9912--9924, 2020.

\bibitem{caron2021emerging}
Mathilde Caron, Hugo Touvron, Ishan Misra, Herv{\'e} J{\'e}gou, Julien Mairal,
  Piotr Bojanowski, and Armand Joulin.
\newblock Emerging properties in self-supervised vision transformers.
\newblock In {\em Proceedings of the IEEE/CVF International Conference on
  Computer Vision}, pages 9650--9660, 2021.

\bibitem{dong2018tri}
Dongdong Chen, Wei Wang, Wei Gao, and Zhi-Hua Zhou.
\newblock Tri-net for semi-supervised deep learning.
\newblock In {\em Proceedings of twenty-seventh international joint conference
  on artificial intelligence}, pages 2014--2020, 2018.

\bibitem{chen2020simple}
Ting Chen, Simon Kornblith, Mohammad Norouzi, and Geoffrey Hinton.
\newblock A simple framework for contrastive learning of visual
  representations.
\newblock In {\em International conference on machine learning}, pages
  1597--1607. PMLR, 2020.

\bibitem{chen2020big}
Ting Chen, Simon Kornblith, Kevin Swersky, Mohammad Norouzi, and Geoffrey~E
  Hinton.
\newblock Big self-supervised models are strong semi-supervised learners.
\newblock {\em Advances in neural information processing systems},
  33:22243--22255, 2020.

\bibitem{chen2020improved}
Xinlei Chen, Haoqi Fan, Ross Girshick, and Kaiming He.
\newblock Improved baselines with momentum contrastive learning.
\newblock {\em arXiv preprint arXiv:2003.04297}, 2020.

\bibitem{chen2021exploring}
Xinlei Chen and Kaiming He.
\newblock Exploring simple siamese representation learning.
\newblock In {\em Proceedings of the IEEE/CVF Conference on Computer Vision and
  Pattern Recognition}, pages 15750--15758, 2021.

\bibitem{chen2020semi}
Yanbei Chen, Xiatian Zhu, Wei Li, and Shaogang Gong.
\newblock Semi-supervised learning under class distribution mismatch.
\newblock In {\em Proceedings of the AAAI Conference on Artificial
  Intelligence}, volume~34, pages 3569--3576, 2020.

\bibitem{cuturi2013sinkhorn}
Marco Cuturi.
\newblock Sinkhorn distances: Lightspeed computation of optimal transport.
\newblock {\em Advances in neural information processing systems}, 26, 2013.

\bibitem{da2022solo}
Victor Guilherme~Turrisi da Costa, Enrico Fini, Moin Nabi, Nicu Sebe, and Elisa
  Ricci.
\newblock solo-learn: A library of self-supervised methods for visual
  representation learning.
\newblock {\em J. Mach. Learn. Res.}, 23:56--1, 2022.

\bibitem{da2022dual}
Victor G~Turrisi da Costa, Giacomo Zara, Paolo Rota, Thiago Oliveira-Santos,
  Nicu Sebe, Vittorio Murino, and Elisa Ricci.
\newblock Dual-head contrastive domain adaptation for video action recognition.
\newblock In {\em Proceedings of the IEEE/CVF Winter Conference on Applications
  of Computer Vision}, pages 1181--1190, 2022.

\bibitem{deng2009imagenet}
Jia Deng, Wei Dong, Richard Socher, Li-Jia Li, Kai Li, and Li Fei-Fei.
\newblock Imagenet: A large-scale hierarchical image database.
\newblock In {\em 2009 IEEE conference on computer vision and pattern
  recognition}, pages 248--255, 2009.

\bibitem{dosovitskiy2021image}
Alexey Dosovitskiy, Lucas Beyer, Alexander Kolesnikov, Dirk Weissenborn,
  Xiaohua Zhai, Thomas Unterthiner, Mostafa Dehghani, Matthias Minderer, Georg
  Heigold, Sylvain Gelly, Jakob Uszkoreit, and Neil Houlsby.
\newblock An image is worth 16x16 words: Transformers for image recognition at
  scale.
\newblock In {\em International Conference on Learning Representations}, 2021.

\bibitem{dwibedi2021little}
Debidatta Dwibedi, Yusuf Aytar, Jonathan Tompson, Pierre Sermanet, and Andrew
  Zisserman.
\newblock With a little help from my friends: Nearest-neighbor contrastive
  learning of visual representations.
\newblock In {\em Proceedings of the IEEE/CVF International Conference on
  Computer Vision}, pages 9588--9597, 2021.

\bibitem{ermolov2021whitening}
Aleksandr Ermolov, Aliaksandr Siarohin, Enver Sangineto, and Nicu Sebe.
\newblock Whitening for self-supervised representation learning.
\newblock In {\em International Conference on Machine Learning}, pages
  3015--3024. PMLR, 2021.

\bibitem{fini2022self}
Enrico Fini, Victor G~Turrisi da Costa, Xavier Alameda-Pineda, Elisa Ricci,
  Karteek Alahari, and Julien Mairal.
\newblock Self-supervised models are continual learners.
\newblock In {\em Proceedings of the IEEE/CVF Conference on Computer Vision and
  Pattern Recognition}, 2022.

\bibitem{fini2021unified}
Enrico Fini, Enver Sangineto, St{\'e}phane Lathuili{\`e}re, Zhun Zhong, Moin
  Nabi, and Elisa Ricci.
\newblock A unified objective for novel class discovery.
\newblock In {\em Proceedings of the IEEE/CVF International Conference on
  Computer Vision}, 2021.

\bibitem{grill2020bootstrap}
Jean-Bastien Grill, Florian Strub, Florent Altch{\'e}, Corentin Tallec, Pierre
  Richemond, Elena Buchatskaya, Carl Doersch, Bernardo Avila~Pires, Zhaohan
  Guo, Mohammad Gheshlaghi~Azar, et~al.
\newblock Bootstrap your own latent-a new approach to self-supervised learning.
\newblock {\em Advances in neural information processing systems},
  33:21271--21284, 2020.

\bibitem{gutmann2010noise}
Michael Gutmann and Aapo Hyv{\"a}rinen.
\newblock Noise-contrastive estimation: A new estimation principle for
  unnormalized statistical models.
\newblock In {\em Proceedings of the thirteenth international conference on
  artificial intelligence and statistics}, pages 297--304, 2010.

\bibitem{he2022masked}
Kaiming He, Xinlei Chen, Saining Xie, Yanghao Li, Piotr Doll{\'a}r, and Ross
  Girshick.
\newblock Masked autoencoders are scalable vision learners.
\newblock In {\em Proceedings of the IEEE/CVF Conference on Computer Vision and
  Pattern Recognition}, pages 16000--16009, 2022.

\bibitem{he2020momentum}
Kaiming He, Haoqi Fan, Yuxin Wu, Saining Xie, and Ross Girshick.
\newblock Momentum contrast for unsupervised visual representation learning.
\newblock In {\em Proceedings of the IEEE/CVF conference on computer vision and
  pattern recognition}, pages 9729--9738, 2020.

\bibitem{he2016deep}
Kaiming He, Xiangyu Zhang, Shaoqing Ren, and Jian Sun.
\newblock Deep residual learning for image recognition.
\newblock In {\em Proceedings of the IEEE conference on computer vision and
  pattern recognition}, pages 770--778, 2016.

\bibitem{huynh2022boosting}
Tri Huynh, Simon Kornblith, Matthew~R Walter, Michael Maire, and Maryam
  Khademi.
\newblock Boosting contrastive self-supervised learning with false negative
  cancellation.
\newblock In {\em Proceedings of the IEEE/CVF Winter Conference on Applications
  of Computer Vision}, pages 2785--2795, 2022.

\bibitem{jaiswal2020survey}
Ashish Jaiswal, Ashwin~Ramesh Babu, Mohammad~Zaki Zadeh, Debapriya Banerjee,
  and Fillia Makedon.
\newblock A survey on contrastive self-supervised learning.
\newblock {\em Technologies}, 9(1):2, 2020.

\bibitem{khosla2020supervised}
Prannay Khosla, Piotr Teterwak, Chen Wang, Aaron Sarna, Yonglong Tian, Phillip
  Isola, Aaron Maschinot, Ce Liu, and Dilip Krishnan.
\newblock Supervised contrastive learning.
\newblock {\em Advances in Neural Information Processing Systems}, 2020.

\bibitem{kim2022conmatch}
Jiwon Kim, Youngjo Min, Daehwan Kim, Gyuseong Lee, Junyoung Seo, Kwangrok Ryoo,
  and Seungryong Kim.
\newblock Conmatch: Semi-supervised learning with confidence-guided consistency
  regularization.
\newblock In {\em European Conference on Computer Vision}, 2022.

\bibitem{krizhevsky2009learning}
Alex Krizhevsky, Geoffrey Hinton, et~al.
\newblock {\em Learning multiple layers of features from tiny images}.
\newblock Toronto, ON, Canada, 2009.

\bibitem{laine2017temporal}
Samuli Laine and Timo Aila.
\newblock Temporal ensembling for semi-supervised learning.
\newblock In {\em International Conference on Learning Representations}, 2017.

\bibitem{lee2013pseudo}
Dong-Hyun Lee et~al.
\newblock Pseudo-label: The simple and efficient semi-supervised learning
  method for deep neural networks.
\newblock In {\em Workshop on challenges in representation learning, ICML},
  2013.

\bibitem{li2021comatch}
Junnan Li, Caiming Xiong, and Steven~CH Hoi.
\newblock Comatch: Semi-supervised learning with contrastive graph
  regularization.
\newblock In {\em Proceedings of the IEEE/CVF International Conference on
  Computer Vision}, pages 9475--9484, 2021.

\bibitem{li2021prototypical}
Junnan Li, Pan Zhou, Caiming Xiong, and Steven Hoi.
\newblock Prototypical contrastive learning of unsupervised representations.
\newblock In {\em International Conference on Learning Representations}, 2021.

\bibitem{liu2022decoupled}
Zicheng Liu, Siyuan Li, Ge Wang, Cheng Tan, Lirong Wu, and Stan~Z Li.
\newblock Decoupled mixup for data-efficient learning.
\newblock {\em arXiv preprint arXiv:2203.10761}, 2022.

\bibitem{mcinnes2018umap}
Leland McInnes, John Healy, and James Melville.
\newblock Umap: Uniform manifold approximation and projection for dimension reduction.
\newblock {\em arXiv preprint arXiv:1802.03426}, 2018.

\bibitem{miyato2016distributional}
Takeru Miyato, Shin-ichi Maeda, Masanori Koyama, Ken Nakae, and Shin Ishii.
\newblock Distributional smoothing with virtual adversarial training.
\newblock In {\em International Conference on Learning Representations}, 2016.

\bibitem{nassar2021all}
Islam Nassar, Samitha Herath, Ehsan Abbasnejad, Wray Buntine, and Gholamreza
  Haffari.
\newblock All labels are not created equal: Enhancing semi-supervision via
  label grouping and co-training.
\newblock In {\em Proceedings of the IEEE/CVF Conference on Computer Vision and
  Pattern Recognition}, pages 7241--7250, 2021.

\bibitem{park2018adversarial}
Sungrae Park, JunKeon Park, Su-Jin Shin, and Il-Chul Moon.
\newblock Adversarial dropout for supervised and semi-supervised learning.
\newblock In {\em Proceedings of the AAAI conference on artificial
  intelligence}, volume~32, 2018.

\bibitem{pham2021meta}
Hieu Pham, Zihang Dai, Qizhe Xie, and Quoc~V Le.
\newblock Meta pseudo labels.
\newblock In {\em Proceedings of the IEEE/CVF Conference on Computer Vision and
  Pattern Recognition}, 2021.

\bibitem{qiao2018deep}
Siyuan Qiao, Wei Shen, Zhishuai Zhang, Bo Wang, and Alan Yuille.
\newblock Deep co-training for semi-supervised image recognition.
\newblock In {\em Proceedings of the european conference on computer vision
  (eccv)}, pages 135--152, 2018.

\bibitem{radford2021learning}
Alec Radford, Jong~Wook Kim, Chris Hallacy, Aditya Ramesh, Gabriel Goh,
  Sandhini Agarwal, Girish Sastry, Amanda Askell, Pamela Mishkin, Jack Clark,
  et~al.
\newblock Learning transferable visual models from natural language
  supervision.
\newblock In {\em International Conference on Machine Learning}, pages
  8748--8763. PMLR, 2021.

\bibitem{sajjadi2016regularization}
Mehdi Sajjadi, Mehran Javanmardi, and Tolga Tasdizen.
\newblock Regularization with stochastic transformations and perturbations for
  deep semi-supervised learning.
\newblock {\em Advances in neural information processing systems}, 29, 2016.

\bibitem{schmutz2022don}
Hugo Schmutz, Olivier Humbert, and Pierre-Alexandre Mattei.
\newblock Don't fear the unlabelled: safe deep semi-supervised learning via
  simple debiaising.
\newblock {\em arXiv preprint arXiv:2203.07512}, 2022.

\bibitem{sohn2020fixmatch}
Kihyuk Sohn, David Berthelot, Nicholas Carlini, Zizhao Zhang, Han Zhang,
  Colin~A Raffel, Ekin~Dogus Cubuk, Alexey Kurakin, and Chun-Liang Li.
\newblock Fixmatch: Simplifying semi-supervised learning with consistency and
  confidence.
\newblock {\em Advances in neural information processing systems}, 33:596--608,
  2020.

\bibitem{szegedy2015going}
Christian Szegedy, Wei Liu, Yangqing Jia, Pierre Sermanet, Scott Reed, Dragomir
  Anguelov, Dumitru Erhan, Vincent Vanhoucke, and Andrew Rabinovich.
\newblock Going deeper with convolutions.
\newblock In {\em Proceedings of the IEEE conference on computer vision and
  pattern recognition}, 2015.

\bibitem{tarvainen2017mean}
Antti Tarvainen and Harri Valpola.
\newblock Mean teachers are better role models: Weight-averaged consistency
  targets improve semi-supervised deep learning results.
\newblock {\em Advances in neural information processing systems}, 30, 2017.

\bibitem{verma2019interpolation}
Vikas Verma, Kenji Kawaguchi, Alex Lamb, Juho Kannala, Yoshua Bengio, and David
  Lopez-Paz.
\newblock Interpolation consistency training for semi-supervised learning.
\newblock In {\em Proceedings of the International Joint Conference on
  Artificial Intelligence, {IJCAI}}, pages 3635--3641, 2019.

\bibitem{vincent2010stacked}
Pascal Vincent, Hugo Larochelle, Isabelle Lajoie, Yoshua Bengio, Pierre-Antoine
  Manzagol, and L{\'e}on Bottou.
\newblock Stacked denoising autoencoders: Learning useful representations in a
  deep network with a local denoising criterion.
\newblock {\em Journal of machine learning research}, 11(12), 2010.

\bibitem{wang2022np}
Jianfeng Wang, Thomas Lukasiewicz, Daniela Massiceti, Xiaolin Hu, Vladimir
  Pavlovic, and Alexandros Neophytou.
\newblock Np-match: When neural processes meet semi-supervised learning.
\newblock In {\em International Conference on Machine Learning}, pages
  22919--22934. PMLR, 2022.

\bibitem{wang2020enaet}
Xiao Wang, Daisuke Kihara, Jiebo Luo, and Guo-Jun Qi.
\newblock Enaet: A self-trained framework for semi-supervised and supervised
  learning with ensemble transformations.
\newblock {\em IEEE Transactions on Image Processing}, 30:1639--1647, 2020.

\bibitem{wang2022debiased}
Xudong Wang, Zhirong Wu, Long Lian, and Stella~X Yu.
\newblock Debiased learning from naturally imbalanced pseudo-labels.
\newblock In {\em Proceedings of the IEEE/CVF Conference on Computer Vision and
  Pattern Recognition}, 2022.

\bibitem{wu2018unsupervised}
Zhirong Wu, Yuanjun Xiong, Stella~X Yu, and Dahua Lin.
\newblock Unsupervised feature learning via non-parametric instance
  discrimination.
\newblock In {\em Proceedings of the IEEE conference on computer vision and
  pattern recognition}, 2018.

\bibitem{xie2020unsupervised}
Qizhe Xie, Zihang Dai, Eduard Hovy, Thang Luong, and Quoc Le.
\newblock Unsupervised data augmentation for consistency training.
\newblock {\em Advances in Neural Information Processing Systems}, 33, 2020.

\bibitem{xie2020self}
Qizhe Xie, Minh-Thang Luong, Eduard Hovy, and Quoc~V Le.
\newblock Self-training with noisy student improves imagenet classification.
\newblock In {\em Proceedings of the IEEE/CVF conference on computer vision and
  pattern recognition}, 2020.

\bibitem{xu2021dash}
Yi Xu, Lei Shang, Jinxing Ye, Qi Qian, Yu-Feng Li, Baigui Sun, Hao Li, and Rong
  Jin.
\newblock Dash: Semi-supervised learning with dynamic thresholding.
\newblock In {\em International Conference on Machine Learning}. PMLR, 2021.

\bibitem{yang2022class}
Fan Yang, Kai Wu, Shuyi Zhang, Guannan Jiang, Yong Liu, Feng Zheng, Wei Zhang,
  Chengjie Wang, and Long Zeng.
\newblock Class-aware contrastive semi-supervised learning.
\newblock In {\em Proceedings of the IEEE/CVF Conference on Computer Vision and
  Pattern Recognition}, pages 14421--14430, 2022.

\bibitem{you2017large}
Yang You, Igor Gitman, and Boris Ginsburg.
\newblock Large batch training of convolutional networks.
\newblock {\em arXiv preprint arXiv:1708.03888}, 2017.

\bibitem{yun2019cutmix}
Sangdoo Yun, Dongyoon Han, Seong~Joon Oh, Sanghyuk Chun, Junsuk Choe, and
  Youngjoon Yoo.
\newblock Cutmix: Regularization strategy to train strong classifiers with
  localizable features.
\newblock In {\em Proceedings of the IEEE/CVF international conference on
  computer vision}, pages 6023--6032, 2019.

\bibitem{zagoruyko2016wide}
Sergey Zagoruyko and Nikos Komodakis.
\newblock Wide residual networks.
\newblock {\em arXiv preprint arXiv:1605.07146}, 2016.

\bibitem{zbontar2021barlow}
Jure Zbontar, Li Jing, Ishan Misra, Yann LeCun, and St{\'e}phane Deny.
\newblock Barlow twins: Self-supervised learning via redundancy reduction.
\newblock In {\em International Conference on Machine Learning}, pages
  12310--12320. PMLR, 2021.

\bibitem{zhai2019s4l}
Xiaohua Zhai, Avital Oliver, Alexander Kolesnikov, and Lucas Beyer.
\newblock S4l: Self-supervised semi-supervised learning.
\newblock In {\em Proceedings of the IEEE/CVF International Conference on
  Computer Vision}, pages 1476--1485, 2019.

\bibitem{zhang2021flexmatch}
Bowen Zhang, Yidong Wang, Wenxin Hou, Hao Wu, Jindong Wang, Manabu Okumura, and
  Takahiro Shinozaki.
\newblock Flexmatch: Boosting semi-supervised learning with curriculum pseudo
  labeling.
\newblock {\em Advances in Neural Information Processing Systems},
  34:18408--18419, 2021.

\bibitem{zhang2018mixup}
Hongyi Zhang, Moustapha Cisse, Yann~N Dauphin, and David Lopez-Paz.
\newblock mixup: Beyond empirical risk minimization.
\newblock In {\em International Conference on Learning Representations}, 2018.

\bibitem{zhang2020wcp}
Liheng Zhang and Guo-Jun Qi.
\newblock Wcp: Worst-case perturbations for semi-supervised deep learning.
\newblock In {\em Proceedings of the IEEE/CVF Conference on Computer Vision and
  Pattern Recognition}, pages 3912--3921, 2020.

\bibitem{zheng2022simmatch}
Mingkai Zheng, Shan You, Lang Huang, Fei Wang, Chen Qian, and Chang Xu.
\newblock Simmatch: Semi-supervised learning with similarity matching.
\newblock In {\em Proceedings of the IEEE/CVF Conference on Computer Vision and
  Pattern Recognition}, pages 14471--14481, 2022.

\bibitem{zhong2021neighborhood}
Zhun Zhong, Enrico Fini, Subhankar Roy, Zhiming Luo, Elisa Ricci, and Nicu
  Sebe.
\newblock Neighborhood contrastive learning for novel class discovery.
\newblock In {\em Proceedings of the IEEE/CVF Conference on Computer Vision and
  Pattern Recognition}, pages 10867--10875, 2021.

\end{thebibliography}
}

\clearpage
\appendix
\normalsize
\section*{Appendix}
\section{More implementation details}
\subsection{ImageNet-1k}
\paragraph{Pre-training.} On ImageNet-1k, we train Suave with ResNet-50 backbone a projection head with hidden and output dimensions 2048 and 128, respectively. The number of prototypes is set to 1000 as the number of classes. We train with mini-batches composed of 256 unlabeled and 128 labeled images. The training lasts for 100 epochs; each epoch consumes all the unlabeled images once. We optimize using LARS with a learning rate of linearly increased from 0 to 0.4 throughout 5 epochs and then decreased to 0.001 with a cosine scheduler. The cross-entropy loss is regularized with a weight decay of $10^{-6}$. Also, the ground-truth labels are smoothed with a factor of 0.01. The pseudo-labels, instead, are computed via three iterations of the Sinkhorn-Knopp algorithm~\cite{cuturi2013sinkhorn} applied to the detached logits (output of the network) extended with a queue of 3840 embeddings buffered from previous mini-batches. The logits used for pseudo labeling are first peaked using a temperature ($\epsilon$ parameter) of 0.05, while the logits used as predictions are peaked with a temperature of 0.1 before computing the loss. On the unlabeled images, we use multi-crop~\cite{caron2020unsupervised} with two large crops (random crop range (0.14, 1)) of size $224^2$ and eight small crops (random crop range (0.08, 0.14)) of size $96^2$. We extend each batch with mixed images generated from MixUp~\cite{yun2019cutmix, zhang2018mixup} with probability 1.0, applying either MixUp or CutMix with probability 0.5 and degree of mixing (known as lambda) drawn from $\mathrm{Beta}(1,1)$. The augmentation recipe of unlabeled images is the exact same as SwAV~\cite{caron2020unsupervised} (color jittering with intensity 0.8 and probability 0.8, random grayscaling with probability 0.2, and Gaussian blurring with probability 0.5). For labeled images, the Inception-style~\cite{szegedy2015going} augmentations adopted consist of random cropping with range (0.08, 1), horizontal flip with probability 0.5, color jittering with intensity 0.4 and probability 0.8, and grayscaling with probability 0.2.

\paragraph{Fine-tuning.} The fine-tuning runs for 3/5 epochs when using 1\%/10\% of the labels with the same semi-supervised setting of the pre-training. Note that the hyper-parameters are kept the same unless specified in the following. The network is fully initialized with the pre-trained weights, except for the prototypes layer, which is randomly initialized. We adopt a smaller learning rate, 0.02, with no linear warm-up and a final value of 0.0002 after cosine decreasing. Also, we reduce the intensity of the augmentations; on the labeled images, we reduce color jittering intensity to 0.1 (keeping probability 0.8) and disable grayscaling; on the unlabeled, we turn off multi-crops, generating only two crops per image with crop range of (0.08, 1), and drop off the color distortions and the blurring.

\paragraph{Simplified training recipe for Daino.} 
For Daino experiments with ViT-S/16 backbone~\cite{dosovitskiy2021image} we adopt the default DINO~\cite{caron2021emerging} pre-training recipe\footnote{see \url{https://dl.fbaipublicfiles.com/dino/dino_deitsmall16_pretrain/args.txt}} for most hyper-parameters except for a few modification that we report in the following. We perform semi-supervised pre-training for 60 epochs, initializing the ViT backbone weights with the DINO pre-trained ones (800 epochs checkpoint); the teacher momentum is set to 0.990; the teacher temperature is raised from 0.04 to 0.07 during the first 10 epochs; the student temperature is fixed to 0.1; we do not freeze the last layer because the labeled loss help to avoid clustering collapse; we set the learning rate to 0.00024 and warm it up linearly for the first 4 epochs; each mini-batch is composed of 1024 unlabeled and 512 labeled images; the labeled images are extended using MixUp~\cite{yun2019cutmix, zhang2018mixup} with probability 1.0 as in the Suave recipe; we augment the unlabeled images with multi-crop obtaining two large crops (crop range 
(0.25,1)) and eight small crops (crop range 
(0.05,0.25)); other data augmentations are maintained as in the original DINO. Note that no fine-tuning is explored for Daino.   

\subsection{CIFAR100}
For CIFAR100 we use a slightly different recipe with respect to ImageNet. First, we do not perform fine-tuning (neither supervised nor semi-supervised), as we found that it does not improve performance. Semi-supervised training is performed with unlabeled batch size 128 and labeled batch size 100 for 200 epochs. For both Suave and Daino, the backbone is initialized using weights obtained by unsupervised training of SwAV for 500 epochs on the same dataset. In addition, we use multi-crop with 4 local crops of size $(0.1, 0.6)$ and 2 global crops of size $(0.6, 1.0)$. Similarly to ImageNet, we use label smoothing with coefficient $0.01$. The learning rate for LARS is set to $2.8$ and a weight decay of $3\cdot10^{-6}$ is applied. The $\epsilon$ coefficients are set to 0.086 and 0.07 for Suave and Daino respectively. For both methods we also use a momentum encoder with momentum $0.99$. We apply image mixing techniques as data augmentation as for ImageNet, with the only difference that we also mix local crops on CIFAR100. All the other hyperparameters are kept the same as described before.

\section{Additional results}
We present further comparisons with the state-of-the-art in Sec.~\ref{ssec:additional_pretrain} and show additional visualizations in Sec.~\ref{ssec:viz}.

\begin{figure*}[ht!]
    \centering
    \begin{subfigure}[b]{.33\textwidth}
        \includegraphics[width=\textwidth]{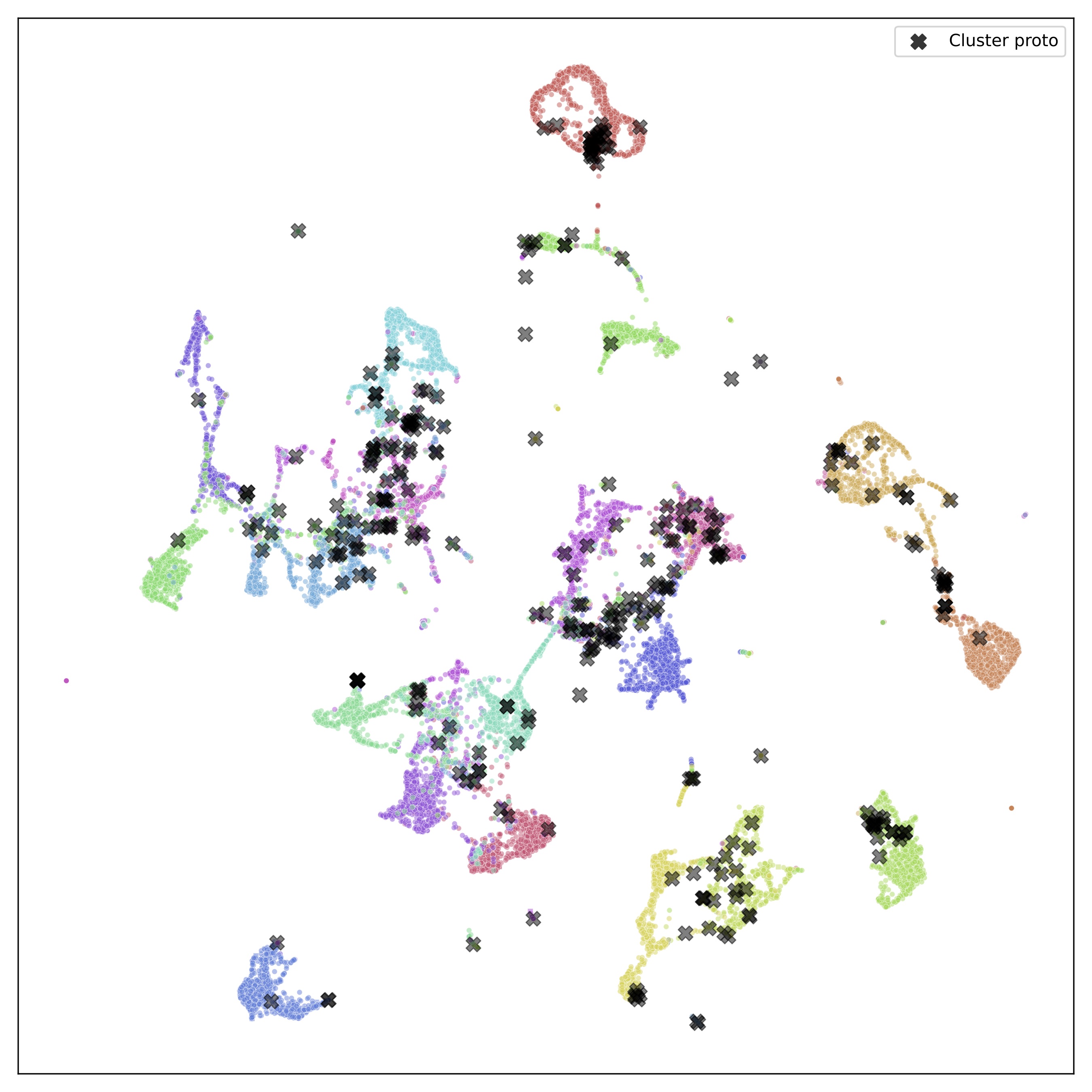}
        \caption{SwAV / SwAV-linear}    
    \end{subfigure}
    \begin{subfigure}[b]{.33\textwidth}
        \includegraphics[width=\textwidth]{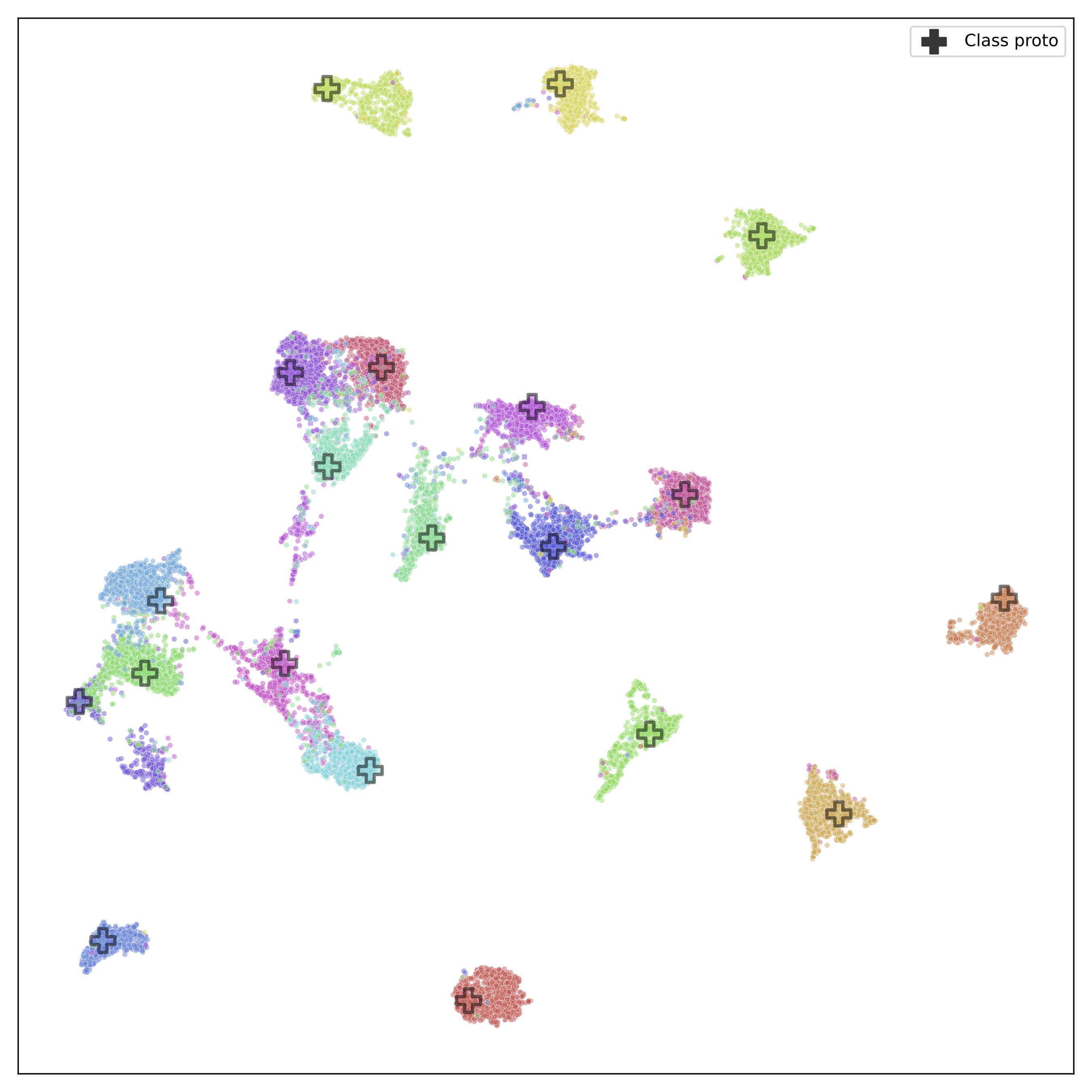}
        \caption{SwAV-finetuned}    
    \end{subfigure}
    \begin{subfigure}[b]{.33\textwidth}
        \includegraphics[width=\textwidth]{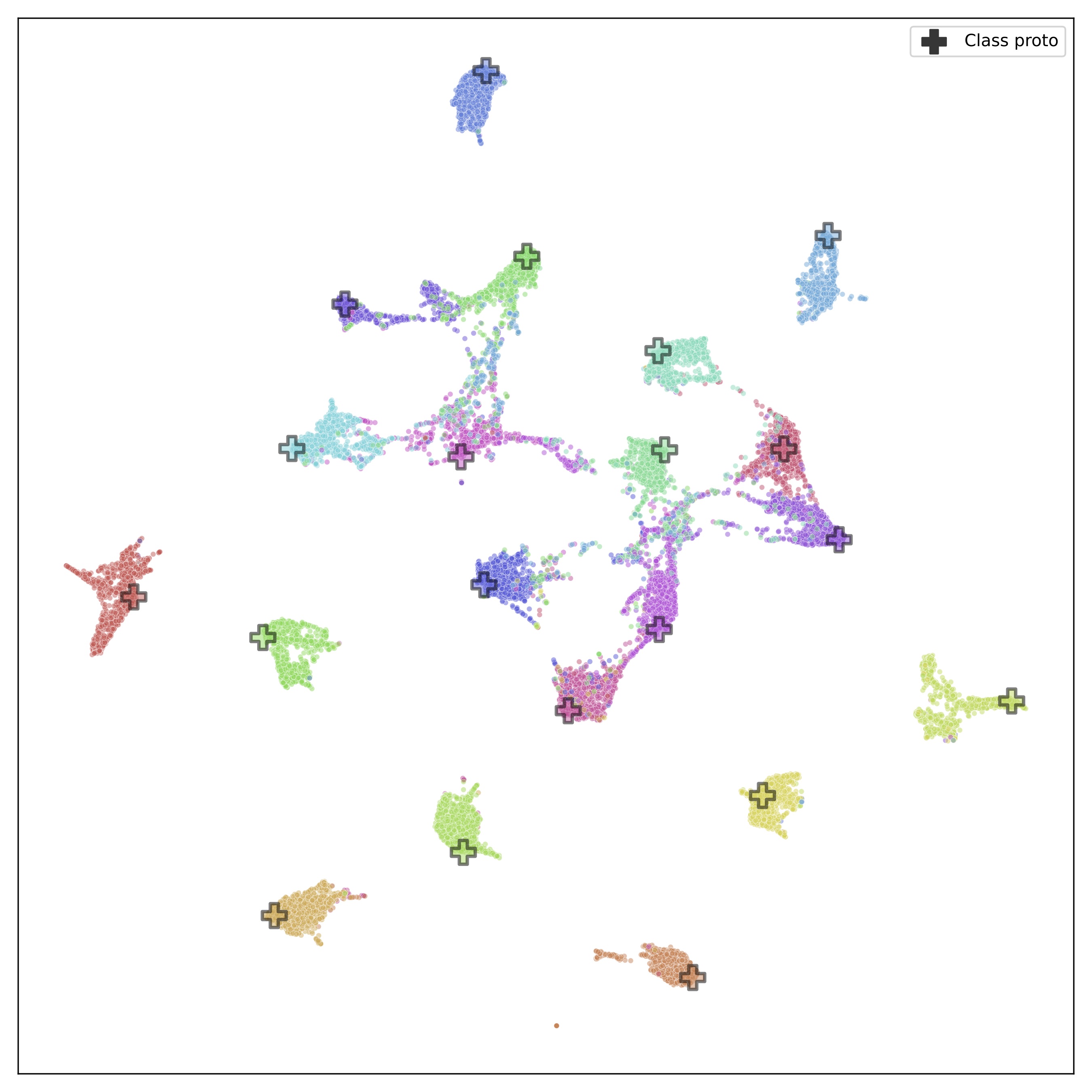}
        \caption{Suave}    
    \end{subfigure}
    \caption{Latent space comparison via UMAP dimensionality reduction. This figure depicts the real-data counterpart of Figure 1 of the main paper. Twenty randomly picked classes are shown and coded by different colors. (a) shows the latent space shared by SwAV and SwAV-linear (which only trains a linear layer for the classification while keeping frozen all the rest of the network). In (a), we also show a random sample of 300 cluster prototypes marked by a red ``X". (b) and (c) show the latent space of SwAV fully fine-tuned and Suave pre-trained and fine-tuned with 1\% of the labels, respectively. Here, the class prototypes are marked by ``+".}
    \label{fig:tsne}
\end{figure*}

\subsection{Pre-training results}
\label{ssec:additional_pretrain}
In Tab.~\ref{suppl:tab:in_results} we report results on ImageNet-1k after semi-supervised pretraining (without fine-tuning) using the same classifier as the one that was trained during pre-training (PAWS uses a nearest-neighbor classifier, we use a linear classifier). The results clearly show that, despite a much smaller batch size, Suave is able to match or outperform PAWS, even without fine-tuning. %

\subsection{Latent representations}
\label{ssec:viz}
In Fig.~\ref{fig:tsne}, we report the real-data counterpart of Figure 1 of the main paper, computed with UMAP~\cite{mcinnes2018umap}. The latent vectors are taken from the bottleneck layer (output of the projection head) of the models trained with 1\% of the labels. All the models are initialized with SwAV pre-trained at 800 epochs.  We observe a neat difference between (a), where classes are less isolated/separable, and (b-c), where, instead, classes are well separated. Moreover, by visually comparing (b) and (c), we notice a slightly better class separation obtained by Suave (c). However, we remark that the random classes shown may not highlight the difference of the models at best, as Suave outperforms SwAV-finetuned of $\sim$9\%p.

\begin{table}[t]
    \centering
    \small
    \caption{Results without fine-tuning on ImageNet-1k.}
    \label{suppl:tab:in_results}
    \resizebox{\linewidth}{!}{%
    \begin{tabular}{lccccc}
    \toprule
    \multirow{2}[1]{*}{\textbf{Method}} & \multirow{2}[1]{*}{\textbf{Epochs}} & \multicolumn{2}{c}{\textbf{Batch size}} & \multicolumn{2}{c}{\textbf{Acc@1}} \\
    \cmidrule(lr){3-4} \cmidrule(lr){5-6}
            &         & Unlab. & Lab.  & 10\% & 1\%  \\ \midrule
    \multirow[t]{3}{*}{PAWS-NN}~\cite{assran2021semi} & 100  & 4096 & 6720 & 71.0 & 61.5 \\
     & 200  & 4096 & 6720 & 71.9 & 63.2 \\ 
     & 300  & 4096 & 6720 & 73.1 & 64.2 \\
    \rowcolor{cyan!20} \multirow[t]{3}{*}{Suave} & (100)100 & 256   & 128  & 71.9 & 62.2 \\
    \rowcolor{cyan!20}    & (200)100  & 256   & 128  & 72.7 & 63.1 \\
    \rowcolor{cyan!20}    & (800)100  & 256   & 128  & 73.4 & 64.1 \\
    \bottomrule
    \end{tabular}%
    }
    \end{table}
\end{document}